\documentclass[11pt, oneside]{article}
\usepackage{geometry}
\usepackage{graphicx}	
\usepackage{amssymb}
\usepackage{amsfonts,latexsym,amsthm,amssymb,amsmath,amscd,euscript}
\usepackage{bbm}
\usepackage{framed}
\usepackage{fullpage}
\usepackage{mathrsfs}
\usepackage{hyperref}
\usepackage{algpseudocode}
\usepackage{algorithm}
\hypersetup{
  colorlinks=true,
  linkcolor=blue,
  filecolor=blue,
  citecolor = black,      
  urlcolor=cyan,
}
\usepackage{accents}
\usepackage{setspace}
\hypersetup{colorlinks=true,citecolor=blue,urlcolor=black,linkbordercolor={1 0 0}}
\usepackage{tikz-cd}

\makeatletter
\newtheorem*{rep@theorem}{\rep@title}
\newcommand{\newreptheorem}[2]{%
\newenvironment{rep#1}[1]{%
 \def\rep@title{#2 \ref{##1}}%
 \begin{rep@theorem}}%
 {\end{rep@theorem}}}
\makeatother

\newcommand{\xref}{\nameref}
\makeatletter
\newcommand\xlabel[2][]{\phantomsection\def\@currentlabelname{#1}\label{#2}}
\makeatother

\theoremstyle{plain}
\newtheorem{theorem}{Theorem}
\newreptheorem{theorem}{Theorem}
\newtheorem{lemma}[theorem]{Lemma}
\newtheorem{corollary}[theorem]{Corollary}
\newtheorem{conjecture}[theorem]{Conjecture}
\newtheorem{proposition}[theorem]{Proposition}

\newtheorem{assumption}[theorem]{Assumption}
\theoremstyle{definition}
\newtheorem{definition}[theorem]{Definition}
\newtheorem{example}[theorem]{Example}

\numberwithin{theorem}{section}

\newcommand{\nc}{\newcommand}
\nc{\DMO}{\DeclareMathOperator}
\newcount\Comments  
\DeclareMathOperator*{\argmin}{arg\,min} 
\DeclareMathOperator*{\argmax}{arg\,max}

\DMO{\prox}{prox}
\DMO{\Span}{span}
\DMO{\UCB}{UCB}
\DMO{\LCB}{LCB}

\nc{\gamvec}{\gamma}
\nc{\til}{\widetilde}
\nc{\td}{\tilde}
\nc{\wh}{\widehat}
\nc{\todo}[1]{\ifnum\Comments=1 {\color{red}  [TODO: #1]}\fi}
\nc{\old}[1]{\ifnum\Comments=1 {\color{brown}  [OLD: #1]}\fi}
\nc{\noah}[1]{\ifnum\Comments=1 {\color{purple} [ng: #1]}\fi}
\nc{\dhruv}[1]{\ifnum\Comments=1 {\color{purple} [dr: #1]}\fi}
\nc{\ankur}[1]{\ifnum\Comments=1 {\color{purple} [am: #1]}\fi}
\nc{\BP}{\mathbb{P}}

\nc{\bel}[1]{\mathbf{b}({#1})}
\nc{\nbel}[1]{\bar\mathbf{b}({#1})}
\nc{\sbel}[2]{\mathbf{b'}_{#1}({#2})}
\nc{\nsbel}[2]{\bar\mathbf{b'}_{#1}({#2})}

\nc{\fools}[3]{\MF_{#3}({#1}, {#2})}
\nc{\fool}[2]{\MF({#1},{#2})}
\nc{\clip}[2]{{\rm clip}\left[ \left. {#1} \right| {#2} \right]}
\nc{\imax}{\omega}
\DMO{\conv}{conv}
\nc{\st}{\star}
\nc{\lng}{\langle}
\nc{\rng}{\rangle}
\DMO{\OOPT}{opt}
\nc{\dopt}[2]{\ell_{\OOPT}({#1},{#2})}
\nc{\grad}{\nabla}
\nc{\MG}{\mathcal{G}}
\nc{\MP}{\mathcal{P}}
\nc{\PP}{\mathbb{P}}
\nc{\TT}{\mathbb{T}}
\nc{\TTmax}{\TT_{\max}}
\DMO{\Reg}{Reg}
\DMO{\Ham}{Ham}
\DMO{\Gap}{Gap}
\DMO{\GD}{GD}
\DMO{\GDA}{GDA}
\DMO{\EG}{EG}
\DMO{\OGDA}{OGDA}
\DMO{\Unif}{Unif}
\DMO{\Tr}{Tr}
\nc{\ul}{\underline}
\nc{\ol}{\overline}
\nc{\Qu}{\ul{Q}}
\nc{\Qo}{\ol{Q}}
\nc{\Ro}{\ol{R}}
\nc{\Vu}{\ul{V}}
\nc{\Vo}{\ol{V}}
\nc{\RanQ}{\Delta Q}
\nc{\RanV}{\Delta V}
\nc{\clipQ}{\Delta \breve{Q}}
\nc{\frzQ}{\Delta \mathring{Q}}
\nc{\clipV}{\Delta \breve{V}}
\nc{\clipdelta}{\breve{\delta}}
\nc{\cliptheta}{\breve{\theta}}
\nc{\delmin}{\Delta_{{\rm min}}}
\nc{\delmins}[1]{\Delta_{{\rm min},{#1}}}
\nc{\gapfinal}[1]{\max \left\{ \frac{\frzQ_{{#1}}^{k^\st}(x,a)}{2H}, \frac{\delmin}{4H} \right\}}
\nc{\post}[2]{R({#1}; {#2})}
\nc{\posts}[3]{R_{#3}({#1}; {#2})}

\nc{\algnst}[1]{\begin{align*}#1\end{align*}}
\nc{\algn}[1]{\begin{align}#1\end{align}}
\nc{\matx}[1]{\left(\begin{matrix}#1\end{matrix}\right)}

\nc{\nuu}{\nu}

\nc{\bv}{\mathbf{v}}
\nc{\bone}{\mathbf{1}}
\nc{\bX}{\mathbf{X}}
\nc{\bY}{\mathbf{Y}}
\nc{\bG}{\mathbf{G}}
\nc{\bz}{\mathbf{z}}
\nc{\bw}{\mathbf{w}}
\nc{\bB}{\mathbf{B}}
\nc{\bA}{\mathbf{A}}
\nc{\bJ}{\mathbf{J}}
\nc{\bK}{\mathbf{K}}
\nc{\bb}{\mathbf{b}}
\nc{\ba}{\mathbf{a}}
\nc{\bc}{\mathbf{c}}
\nc{\bC}{\mathbf{C}}
\nc{\BR}{\mathbb R}
\nc{\BA}{\mathbb{A}}
\nc{\BC}{\mathbb C}
\nc{\bx}{\mathbf{x}}
\nc{\bS}{\mathbf{S}}
\nc{\bM}{\mathbf{M}}
\nc{\bR}{\mathbf{R}}
\nc{\bN}{\mathbf{N}}
\nc{\by}{\mathbf{y}}
\nc{\sy}{y}
\nc{\sx}{x}

\nc{\MO}{\mathcal O}
\nc{\MU}{\mathcal{U}}
\nc{\ME}{\mathcal{E}}
\nc{\MN}{\mathcal{N}}
\nc{\MK}{\mathcal{K}}
\nc{\MM}{\mathcal{M}}
\nc{\MS}{\mathcal{S}}
\nc{\MT}{\mathcal{T}}
\nc{\BF}{\mathbb F}
\nc{\BQ}{\mathbb Q}
\nc{\MX}{\mathcal{X}}
\nc{\MA}{\mathcal{A}}
\nc{\MD}{\mathcal{D}}
\nc{\MB}{\mathcal{B}}
\nc{\MZ}{\mathcal{Z}}
\nc{\MJ}{\mathcal{J}}
\nc{\MW}{\mathcal{W}}
\nc{\MR}{\mathcal{R}}
\nc{\MY}{\mathcal{Y}}
\nc{\BZ}{\mathbb Z}
\nc{\BN}{\mathbb N}
\nc{\ep}{\epsilon}
\nc{\gapfn}[1]{\varepsilon_{#1}}
\nc{\ggapfn}[2]{\varphi_{#1}({#2})}
\nc{\epsahk}{\gapfn{0}}
\nc{\BH}{\mathbb H}
\nc{\BG}{\mathbb{G}}
\nc{\D}{\Delta}
\nc{\MF}{\mathcal{F}}
\nc{\One}{\mathbbm{1}}
\nc{\bOne}{\mathbf{1}}
\nc{\Aopt}{\mathcal{A}^{\rm opt}}
\nc{\Amul}{\mathcal{A}^{\rm mul}}

\nc{\SP}{\mathsf P}
\nc{\SQ}{\mathsf Q}

\nc{\DO}{\accentset{\circ}{\D}}
\nc{\mf}{\mathfrak}
\nc{\mfp}{\mathfrak{p}}
\nc{\mfq}{\mf{q}}
\nc{\Sp}{\mbox{Spec}}
\nc{\Spm}{\mbox{Specm}}
\nc{\hookuparrow}{\mathrel{\rotatebox[origin=c]{90}{$\hookrightarrow$}}}
\nc{\hookdownarrow}{\mathrel{\rotatebox[origin=c]{-90}{$\hookrightarrow$}}}
\nc{\hra}{\hookrightarrow}
\nc{\tra}{\twoheadrightarrow}
\nc{\sgn}{{\rm sgn}}
\nc{\aut}{{\rm Aut}}
\nc{\Hom}{{\rm Hom}}
\nc{\img}{{\rm Im}}
\DMO{\id}{Id}
\DMO{\supp}{supp}
\DMO{\KL}{KL}
\nc{\kld}[2]{\KL({#1}||{#2})}
\nc{\tvd}[2]{\left\| {#1} - {#2} \right\|_1}
\nc{\ren}[2]{D_2({#1}||{#2})}
\nc{\chisq}[2]{\chi^2({#1}||{#2})}
\nc{\hell}[2]{H^2({#1}, {#2})}
\nc{\E}{\mathbb{E}}

\DMO{\BSS}{BSS}
\DMO{\BES}{BES}
\DMO{\BGS}{BGS}
\DMO{\poly}{poly}
\nc{\indep}{\perp}
\DMO{\sink}{sink}
\nc{\fp}[1]{\MP_1({#1})}

\nc{\RR}{\mathbb{R}}
\nc{\Gradient}{\nabla}
\DMO{\diag}{diag}
\nc{\norm}[1]{\left \lVert #1 \right \rVert}
\nc{\bO}{\mathbb{O}}
\nc{\bT}{\mathbb{T}}
\nc{\BO}{\bO}
\nc{\BT}{\bT}
\nc{\CH}{\mathscr{H}}
\DMO{\TV}{TV}
\nc{\PSPACE}{\textsf{PSPACE}}
\nc{\ETH}{\textsf{ETH}}
\nc{\NP}{\textsf{NP}}
\nc{\MerlinArthur}{\textsf{MA}}
\nc{\AM}{\textsf{AM}}
\nc{\coNP}{\textsf{coNP}}
\nc{\EH}{\textsf{EH}}

\DMO{\PR}{Pr}
\renewcommand{\Pr}{\PR}
\DeclareMathOperator*{\EE}{\mathbb{E}}
\nc{\ra}{\rightarrow}
\renewcommand{\t}{\top}

\nc{\belief}{\textbf{b}}
\nc{\sbelief}{\textbf{b}'}

\nc{\Renyi}{\text{R\'enyi }}

\title{Planning in Observable POMDPs in Quasipolynomial Time}

\author{Noah Golowich\thanks{Email: \texttt{nzg@mit.edu}. Supported by a Fannie \& John Hertz Foundation Fellowship and an NSF Graduate Fellowship.} \\ MIT \and Ankur Moitra\thanks{Email: \texttt{moitra@mit.edu}. Supported in part by a Microsoft Trustworthy AI Grant, NSF Large CCF1565235, NSF CCF1918421 and a David and Lucile Packard Fellowship.} \\ MIT \and Dhruv Rohatgi\thanks{Email: \texttt{drohatgi@mit.edu}. Supported by an Akamai Presidential Fellowship and a U.S. DoD NDSEG Fellowship.} \\ MIT}
\date{\today}

\begin{document}
\maketitle

\begin{abstract}
Partially Observable Markov Decision Processes (POMDPs) are a natural and general model in reinforcement learning that take into account the agent's uncertainty about its current state. In the literature on POMDPs, it is customary to assume access to a planning oracle that computes an optimal policy when the parameters are known, even though the problem is known to be computationally hard. Almost all existing planning algorithms either run in exponential time, lack provable performance guarantees, or require placing strong assumptions on the transition dynamics under every possible policy. In this work, we revisit the planning problem and ask: {\em Are there natural and well-motivated assumptions that make planning easy?}

Our main result is a quasipolynomial-time algorithm for planning in (one-step) observable POMDPs. Specifically, we assume that well-separated distributions on states lead to well-separated distributions on observations, and thus the observations are at least somewhat informative in each step. Crucially, this assumption places no restrictions on the transition dynamics of the POMDP; nevertheless, it implies that near-optimal policies admit quasi-succinct descriptions, which is not true in general (under standard hardness assumptions). Our analysis is based on new quantitative bounds for filter stability \--- i.e. the rate at which an optimal filter for the latent state forgets its initialization. Furthermore, we prove matching hardness for planning in observable POMDPs under the Exponential Time Hypothesis. 

\end{abstract}

\newpage 

\section{Introduction}
 A key challenge in reinforcement learning (RL) is that we often cannot fully observe the state of the system. Instead, we have access to noisy and/or incomplete measurements. Such situations arise in the application of RL in robotics, where sensors may not be able to capture all data about a robot's configuration \cite{cassandra1996acting}, in game theory, where agents cannot observe others' private information \cite{brown2018superhuman}, in healthcare, where clinical measurements do not capture a patient's full physiological state \cite{hauskrecht2000planning,futoma2020popcorn}, and in many other settings. The standard approach is to model interactions with the system through a \emph{partially observable Markov decision process (POMDP)}, which generalizes a (fully observed) \emph{Markov decision process (MDP)} by allowing the learning agent to see only a noisy observation that depends on the hidden state. 
In addition to being a more realistic model for learning in sequential decision making environments, POMDPs are surprisingly powerful, having the capacity to model common broad problem frameworks within RL such as meta RL and robust RL \cite{ni2021recurrent}.

Despite their broad importance, there is a wide gap between theory and practice when it comes to our understanding of POMDPs. The most basic problem we could ask to solve is that of finding a (nearly) optimal policy (i.e., a strategy for how to interact with the environment) when the model is fully known. This problem is known as \emph{(approximate) planning in POMDPs}. In contrast to planning in MDPs, planning in POMDPs is computationally intractable in general, and most theoretical results are negative. Intuitively, the central difficulty is that POMDPs lack the Markovian property of MDPs: the optimal action to take at some step may depend not just on the most recent observation but on the entire history of actions and observations. This has been dubbed the \emph{Curse of History} \cite{pineau2006anytime}: an optimal policy could take exponential space even to describe, since the number of possible interaction histories depends exponentially on the horizon. Concretely, the seminal work of Papadimitriou and Tsitsiklis \cite{papadimitriou1987complexity} showed that computing an optimal policy is \PSPACE-hard. However in the setting where the rewards are bounded in $[0,1]$, their result only provides weak quantitative lower bounds in the sense that they show hardness for approximating the optimal policy to within exponentially small additive error. Later works used similar reductions to show that even for constant $\epsilon$, computing an $\epsilon$-additive approximation to the optimal policy value is \PSPACE-hard \cite{littman1994memoryless, burago1996complexity, lusena2001nonapproximability}. 

On the practical side, a variety of heuristics for approximate planning in POMDPs have been proposed \cite{jaakkola1995reinforcement, hansen1998solving, hauskrecht2000value, roy2002exponential, theocharous2003approximate, poupart2004vdcbpi, spaan2005perseus, pineau2006anytime, ross2008online, silver2010monte, smith2012heuristic, somani2013despot, garg2019despot}, and have found some success in circumventing the worst-case intractability. A popular approach is to discretize the space of \emph{beliefs} we maintain about the hidden state. This trades off the Curse of History for the {\em Curse of Dimensionality}, since the size of the discretization needs to be exponential in the number of hidden states \cite{kaelbling1998planning, pineau2006anytime}. Another approach is to optimize the policy over a constrained class of functions \cite{jaakkola1995reinforcement, hansen1998solving} \--- for example, \emph{memoryless} policies, which only depend on the last observation. Such policies can be succinctly described, but may have arbitrarily poor performance compared to the optimal policy. This latter approach is also common in deep RL, where the policies are parametrized by a neural network \cite{hausknecht2015deep}. While some of these approaches have obtained promising empirical results, there is still much room for improvement since at best they converge to a locally optimal, but globally suboptimal, policy. Even worse, the problem of finding the best memoryless policy is \NP-hard \cite{littman1994memoryless}. 

Beyond the importance of the problem of planning in POMDPs in its own right, it is also a key primitive in the learning problem, where the goal is to learn a good policy in an \emph{unknown} environment. Obviously, this problem inherits the computational woes of planning. Thus it is standard to simply ignore computational issues, and focus instead on issues of exploration and sample-efficiency, namely the number of episodes of interaction needed to learn a good policy. In particular, the convention in the recent theoretical RL literature on learning POMDPs \cite{krishnamurthy2016pac, guo2016pac, azizzadenesheli2016reinforcement, jin2020sample, xiong2021sublinear, kwon2021reinforcement, kwon2021rl} is to assume access to an \emph{oracle} that solves the POMDP planning problem. There are some exceptions, but they require very strong assumptions on the model, which essentially trivialize the planning aspect \--- for instance, they assume that the state transitions are deterministic \cite{krishnamurthy2016pac, jin2020sample}, in which case (for known model) we always know the hidden state even without receiving any observations, or they assume that the ``unobserved'' portion of the state has constant size and never changes \cite{kwon2021reinforcement}.

Surveying three decades of research on POMDPs therefore gives the impression that the story of approximate planning is closed (at least, for theoreticians). There are formidable obstacles to reconciling theory and practice. Not only is the algorithmic problem of finding a good policy computationally hard, but the problem of succinctly describing it is likely impossible: Papadimitriou and Tsitsiklis \cite{papadimitriou1987complexity} also proved that unless $\PSPACE = \Sigma_2^P$, optimal policies cannot be described with polynomial space. In Appendix~\ref{app:succinct-policies} we extend their results and show that unless the exponential hierarchy collapses, there is no quasipolynomial space description even for an approximately optimal policy. 

However, these are \emph{worst-case} computational intractability results. 
Indeed there are natural assumptions in the literature on learning in POMDPs that are sufficient for avoiding worst-case \emph{sample complexity} lower bounds \cite{jin2020sample}. Thus, we can hope to bring similar perspectives to the problem of computationally efficient planning \--- where natural and well-motivated assumptions about POMDPs are lacking \--- with the ultimate aim of proving strong end-to-end algorithmic guarantees for rich classes of problems.


\paragraph{Observability.} In this paper, we work with the following assumption, which we refer to as \emph{observability}, and was introduced by \cite{even2007value} (under the name ``value of observation'') for a different but complementary goal of understanding the stability of beliefs in Hidden Markov Models (HMMs) when the parameters are misspecified. Let $H \in \BN$ denote the horizon of the POMDP, and for each step $h \in [H]$, let $\bO_h(\cdot | x_h)$ denote the distribution over observations given that the true (latent) state is $x_h$ at step $h$. Observability is then formally defined as follows:

\begin{assumption}[\cite{even2007value}]\label{assumption:observability-intro}
  Let $\gamma > 0$. For $h \in [H]$, let $\bO_h$ be the matrix with rows $\bO_h(\cdot|x_h)$. We say that the matrix $\bO_h$ satisfies \emph{$\gamma$-observability} if for each $h$, for any distributions $b,b'$ over states, $$\norm{\bO_h^\t b - \bO_h^\t b'}_1 \geq \gamma \norm{b - b'}_1.$$
  A POMDP (or HMM) satisfies \emph{(one-step) $\gamma$-observability} if all $H$ of its observation matrices do. 
\end{assumption}

Intuitively, this assumption stipulates that the observations are at least somewhat informative. In particular, well-separated distributions over states lead to well-separated distributions over observations. One example of an observation matrix which satisfies $\gamma$-observability with $\gamma = 1/2$ is the random channel which outputs the hidden state with probability $1/2$ and otherwise outputs a random state (i.e. from a ``noisy sensor''). Even planning under this specific but natural observation model was to our knowledge unresolved until the present work. This condition has also been invoked in the context of collaborative filtering \cite{kleinberg2008using} and inference in topic models \cite{arora2016provable} for similar reasons. 

Why is this assumption reasonable? The crucial point is that it makes no restriction at all on the model's transitions. As noted by \cite{even2007value}, transition dynamics are often an immutable property of the system. In contrast, observability may in practice be more directly under our control since we may be able to, for instance, decrease the noise level in the observations by placing more or better sensors. The distinction between making assumptions on the dynamics and on the observations is even more acute in controlled systems such as POMDPs, since the transitions depend on the agent's actions, and it is usually unreasonable to make assumptions (such as mixing or ergodicity) that must hold for all policies.

\paragraph{Main results.} In this paper, we prove that for any constant $\gamma > 0$, the $\gamma$-observability assumption enables quasi-succinct descriptions of near-optimal policies. Moreover, we give a quasi-polynomial time approximate planning algorithm for POMDPs. {\em To our knowledge, this is the first provable approximate planning algorithm for POMDPs that does not take exponential time, makes no assumptions about the model transitions, and allows for a broad family of observation models.}

\begin{theorem}
  \label{thm:main-intro}
Let $\epsilon,\gamma > 0$. There is an algorithm which given the description of a $\gamma$-observable POMDP with $S$ states, $A$ actions, $O$ observations, and horizon length $H$, outputs an $\epsilon$-suboptimal policy and has time complexity $H(OA)^{C\log(SH/\epsilon)/\gamma^4}$ for some universal constant $C>0$.
\end{theorem}

The key to this theorem is a much stronger notion of belief \emph{stability} compared to the results in \cite{even2007value}. First, note that if we fix a policy, the hidden state and observations follow an HMM. For any prior (i.e. belief) on the latent state at some time, the Bayes filter is a way of computing the induced posterior on the latent state at any future time. If the prior is incorrect, then the posterior will also be incorrect; however, one might hope that the filter is ``stable'' in the sense that the error will decay over time. Indeed, the techniques in \cite{even2007value} show that for observable HMMs, initialization error of an approximate belief state decays polynomially fast under the Bayes filter in expectation, assuming that the approximate belief state is initialized to the uniform distribution. 

This is not strong enough to obtain a subexponential-time planning algorithm for observable POMDPs, because if we restricted our policy to depend on only a window of length $\ell$ of the most recent observations and actions, we would have to take $\ell$ to be polynomial in $H$ in order to get sufficient approximation guarantees. Instead, we show that under the same assumptions, the Bayes filter is in fact \emph{exponentially stable} in expectation in the following sense: 

\begin{theorem}[Exponential stability of belief states; informal statement of Theorem \ref{theorem:stability-renyi}]
  \label{theorem:stability-informal}
For any $\gamma$-observable POMDP with state space size $S$ and horizon length $H$, fix a policy $\pi$ and timesteps $h < h+t \leq H$. Let $b_h$ be an arbitrary belief state, and let $b'_h$ be the uniform belief state. Pick an initial state $x_h \sim b_h$ and draw a trajectory $\tau$ from the POMDP by following policy $\pi$. If $b_{h+t}$ and $b'_{h+t}$ are the posteriors for state $x_{h+t}$ under trajectory $\tau$ and priors $b_h$ and $b'_h$, respectively, then $$\EE_\tau \left[\norm{b_{h+t} - b'_{h+t}}_1 \right]\leq (1 - \gamma^4)^t S.$$
\end{theorem}

\noindent Thus the expected total variation distance between beliefs is at most $\epsilon$ after only $\widetilde{O}(\gamma^{-4}\log(1/\epsilon))$ steps, where the $\widetilde O$ hides polynomial dependence on $\log S$. With a related argument, we can also show that $\widetilde{O}(\gamma^{-2}\log^2(1/\epsilon))$ steps also suffice (Theorem~\ref{theorem:stability-kl}, which is used to obtain a corresponding time complexity of $H(OA)^{O(\log^2(SH/\ep)/\gamma^2)}$ in Theorem \ref{thm:main-kl}). This bound of Theorem \ref{theorem:stability-kl} is incomparable, but as we show in Proposition~\ref{prop:belief-contraction-lower-bound}, its dependence on $\gamma$ is optimal.

Theorem~\ref{theorem:stability-informal} is also of independent interest; 
it can be seen as bringing filter stability results for finite HMMs in line with those for infinite but highly structured HMMs. It is well-known that analogues of observability lead to exponential stability for structured HMMs such as linear dynamical systems (which have infinite state spaces but very rigid transition rules) \cite{moore1980coping,schick1989robust, atar1997lyapunov, atar1997exponential}. Moreover, this property plays a key role in applications, in terms of bounding the effect of misspecification and analyzing the asymptotic \cite{mitter1992point,schick1994robust} and non-asymptotic \cite{chen2021kalman} robustness. For finite, unstructured HMMs (our setting), there are asymptotic stability results under analogues of observability \cite{van2009observability}, and exponential stability results under mixing assumptions \cite{shue1998exponential, boyen2013tractable, mcdonald2020exponential}. In this context, Theorem~\ref{theorem:stability-informal} is the first exponential stability result for unstructured HMMs \emph{without} mixing assumptions. See Section~\ref{section:related-work} for a more detailed discussion of the prior work.

\paragraph{Lower bounds.} We next show that the upper bound of Theorem \ref{thm:main-intro} cannot be substantially improved: in particular, there is no polynomial-time algorithm for planning in observable POMDPs even for constant $\gamma$, and furthermore an inverse-polynomial dependence of the exponent on $\gamma$ is necessary.
\begin{theorem}[Lower bound; informal statement of Theorem \ref{thm:eth-lb}]\label{theorem:lb-intro}
Under the Exponential Time Hypothesis, no algorithm for planning in $\gamma$-observable POMDPs running in time $(SAHO)^{o(\log(SAHO/\epsilon)/\gamma)}$  can produce $\epsilon$-suboptimal policies. 
\end{theorem}

It is natural to ask whether the $\gamma$-observability assumption can be weakened to a condition on pairs of states rather than on pairs of distributions on states. In particular, ``weak observability" stipulates that for any two states $x,x'$, it holds that $\norm{\bO_h(\cdot|x) - \bO_h(\cdot|x')}_1 \geq \gamma$; this assumption is used for learning latent MDPs in \cite{kwon2021rl} (although even in that restricted setting, a planning oracle is used). Unfortunately, we provide a simple example showing that planning in weakly observable POMDPs is still \ETH-hard. 

\begin{proposition}[Lower bound against weak observability]\label{prop:weak-observability-lower-bound}
Under the Exponential Time Hypothesis, for some absolute constant $c>0$, no algorithm for planning in $1/2$-weakly-observable POMDPs can produce $\epsilon$-suboptimal policies in time $2^{O((SAOH/\epsilon)^c)}$.
\end{proposition}

\paragraph{Discussion.} Our work is motivated by a careful examination and critique of the lower bound instances in \cite{papadimitriou1987complexity}. In particular, the instances that arise from their (simpler, \NP-hardness) reduction have the property that we get {\em no} observations about the hidden state. Thus, the POMDPs are not just not fully observable, but in fact are completely unobservable. Observability is a natural assumption that circumvents this lower bound and in fact, as we show, leads to a more efficient planning algorithm. Nevertheless we view observability as just a first step in unraveling classes of reasonable assumptions that allow us to prove positive results. Towards that end, we ask: Are there multi-step generalizations of observability that allow for positive results? In a different direction, we remark that Assumption \ref{assumption:observability-intro} can only hold in the \emph{undercomplete} setting, namely where there are at least as many observations as states. It is interesting to investigate whether there are natural assumptions under which one can establish provable guarantees for planning in \emph{overcomplete} POMDPs. 

\subsection{Related work}\label{section:related-work}

\paragraph{Planning algorithms for POMDPs.} There is a large literature on planning in POMDPs. However, almost all such algorithms require exponential time in the number of states or horizon length \cite{sondik1978optimal, kaelbling1998planning, pineau2006anytime, mcallester2013approximate}, or lack provable performance guarantees with respect to the optimal policy. Examples of the latter type include algorithms employing various pruning heuristics \cite{monahan1982state,cassandra1997incremental,hansen2000dynamic,hauskrecht2000value, roy2002exponential, theocharous2003approximate, poupart2004vdcbpi, spaan2005perseus, pineau2006anytime, ross2008online, silver2010monte, smith2012heuristic, somani2013despot, garg2019despot} and algorithms which optimize over restricted classes of policies \cite{hansen1998solving, meuleau1999solving,kearns1999approximate, li2011finding, azizzadenesheli2018policy}. To our knowledge, some of the only works presenting sub-exponential time approximate planning algorithms are \cite{burago1996complexity} and \cite{mcdonald2020exponential,kara2020near} (ignoring end-to-end learning algorithms, which we discuss later).

Most relevant to our work is the sequence of papers \cite{mcdonald2020exponential, kara2020near}, which shows in similar vein to our work that under certain assumptions about the POMDP, a short-memory policy is nearly optimal. The key difference is that their work makes strong mixing assumptions about the transition matrix, which must hold for every action. In contrast, our Assumption~\ref{assumption:observability-intro} only involves the observations. 
In \cite{burago1996complexity}, a polynomial-time approximate planning algorithm is given for POMDPs where the observations are deterministic and every observation is generated by at most a constant number of states. To our knowledge, this is one of the only sub-exponential time planning algorithms for POMDPs which, like our result, needs no assumptions about the transition matrices. However, the assumption on the observation model is rather stringent, in that it does not allow for stochasticity in the observations (e.g. low-probability failure events in which the observation is completely uninformative).

Finally, while our focus in this paper is on the finite horizon setting, we remark that in the related infinite-horizon discounted setting for POMDPs, determining whether there exists a policy with value at least some given threshold is \emph{undecidable} \cite{madani2003undecidability}.








\paragraph{Learning algorithms for POMDPs.} There is also a large literature on learning in POMDPs \cite{even2005reinforcement, jin2020sample, xiong2021sublinear, kwon2021rl, kwon2021reinforcement, guo2016pac, azizzadenesheli2016reinforcement, krishnamurthy2016pac, azizzadenesheli2018policy}. Almost all are either exponential-complexity in the worst case (e.g. \cite{even2005reinforcement}) or make use of a planning oracle (or an ``optimistic planning'' oracle which seems even more intractable, e.g. \cite{jin2020sample}), or implement the planning oracle but via an exponential-time algorithm (e.g. \cite{guo2016pac}). Below we discuss the exceptions, and the assumptions that they require.

First, the main learning result of \cite{jin2020sample} uses an optimization oracle which is likely intractable. However, a secondary result of theirs is a computationally efficient learning algorithm for POMDPs with deterministic transitions and deterministic initial state, as well as inverse-polynomial distance between observation distributions of any two states. Compared to our regime, the assumption on the observation model is much weaker, but the deterministic transition assumption is strong; in fact, the planning algorithm in this setting is trivial, because there is no uncertainty in the state. The learning algorithm of \cite{krishnamurthy2016pac} operates in the function approximation setting. Here, the transitions and initial state are both assumed to be deterministic. Additionally, it is assumed that the optimal policy and optimal $Q$-function are memoryless. 

Finally, \cite{kwon2021reinforcement} provides a computationally efficient learning algorithm for a special case of POMDPs which they call reward-mixing MDPs. Specifically, the POMDP is a disjoint union of two MDPs which are identical except for the rewards. The initial state distribution is arbitrary across the disjoint union, and the observation at state $s$ in MDP $m$ is simply $s$ (i.e. the choice of MDP is never observed). Here, once again, planning is straightforward; the belief space is $\MS \times [0,1]$, which has (covering) dimension $1$, so classical exponential-time approaches such as belief discretization are computationally efficient in this setting.










\paragraph{Filter stability.} Filter stability is a well-studied topic in stochastic processes and dynamical systems. Formally, for a latent process $(X_t)_t$ with observations $(Y_t)_t$ (i.e. a Hidden Markov Model), a \emph{filter} is an evolving estimate of the latent state $X_t$ based on the observations $Y_{1:t}$, which can be recursively computed from $Y_t$ and the estimator for $X_{t-1}$. When the model is fully specified, the optimal filter (in certain senses) is the Bayes filter, which at time $t$ is defined as the posterior distribution of $X_t$ given $Y_{1:t}$. It can be efficiently computed for finite HMMs as well as certain well-structured infinite HMMs---for instance, in linear dynamical systems with Gaussian process and observation noise, it reduces to the Kalman filter.

\emph{Filter stability} is the question of whether and how quickly the filter will ``forget'' an incorrect initialization. This depends in general on both the transition dynamics and the observation model. In broad strokes, there are two classes of known filter stability results. On the one hand, there is a rich literature on filter stability for structured infinite HMMs, particularly linear dynamical systems and processes with additive Gaussian noise \cite{schick1989robust, schick1994robust, atar1997lyapunov, atar1997exponential}. These results achieve exponential filter stability under analogues of our observability assumption in addition to other assumptions such as controllability. On the other hand, closer to our setting, there is work on filter stability for unstructured but finite or compact HMMs.

In an asymptotic sense, the latter problem is essentially solved. The Bayes filter for a finite HMM is asymptotically stable if and only if it is \emph{detectable}, meaning that for any two initial latent distributions $\mu, \nu$ which induce the same observation process, the induced latent process distributions merge over time \cite{van2009observability}. In particular, to make explicit the connection with our work, a sufficient condition for detectability is \emph{uniform observability} \cite{van2009observability,vanhandel2009uniform}, which in stochastic process theory refers to the assumption that for any pair of latent distributions $\mu,\nu$, if the induced observation processes are sufficiently close, then $\mu$ and $\nu$ must be close as well. The reason why uniform observability suffices for asymptotic stability is a classical result by Blackwell and Dubins \cite{blackwell1962merging} that (with no assumptions other than absolute continuity of the filter's initialization with respect to the truth) the filter's prediction of $Y_{t+1:\infty}$ at time $t$ must merge with the true distribution as $t \to \infty$. This connection has been observed several times \cite{chigansky2006role, van2009observability, mcdonald2018stability}; see \cite{van2010nonlinear} for an exposition.

\paragraph{Non-asymptotic bounds?}
Unfortunately, the uniform observability assumption is inherently asymptotic \--- it only requires that the infinite sequence of observations is informative. In contrast, we require that each observation by itself is informative. Moreover, even if we make a quantitative assumption about the value of each observation, the proof technique of combining the unconditional result of Blackwell and Dubins with observability is fundamentally insufficient for exponential stability: while Blackwell and Dubins' result is asymptotic and can be made quantitative with polynomial convergence \cite{sharan2018prediction} in the setting with a finite state space, such polynomial convergence cannot be improved (see Corollary~\ref{cor:blackwell-dubins-converse} or \cite{sharan2018prediction}).

There are several works on exponential stability for unstructured HMMs \cite{shue1998exponential, boyen2013tractable, mcdonald2018stability}. However, these make strong mixing assumptions about the transitions. To give an example of why these are unsatisfactory, \cite{shue1998exponential} assumes that all entries of all transition matrices are strictly positive; the proof is then via the Perron-Frobenius theorem. Thus, to get any reasonable quantitative bounds it is necessary to assume that each state is reachable from every other state in a small number of steps. To give another example, \cite{mcdonald2018stability} assumes a bound on the Dobrushin coefficient of the transitions, which is violated if there are two states and one action such that the induced transition distributions are disjoint. In fact, \cite{mcdonald2018stability} also assumes a similar bound on the observations, which is intuitively the \emph{opposite} of observability: the bound gets worse as the observations become more informative. 

Aside from \cite{even2005reinforcement}, which achieves polynomial stability under observability, we are aware of no results with {\em quantitative} bounds on filter stability (let alone exponential stability) for unstructured but finite HMMs, that fundamentally use the value of observations and do not require assumptions on the transitions.

\paragraph{Prediction with short memory.} Recent work by Sharan, Kadade, Liang, and Valiant \cite{sharan2018prediction} has a similar message to ours: that for finite HMMs, only a short suffix of the history needs to be remembered to make nearly accurate predictions. However, these are predictions of the \emph{observations}, not the latent state, and no observability assumption is made. As a result, their work is more closely related to Blackwell and Dubins' ``merging of opinions'' \cite{blackwell1962merging}. They prove inverse polynomial stability of the observation distributions, which together with observability does imply inverse polynomial stability of the latent distributions. However, as previously discussed, their rate is optimal, which is why we need a different approach to prove exponential stability of the latent distributions under observability.




\subsection{Organization}

In Section~\ref{section:preliminaries}, we introduce notation, including a formal description of the POMDP model. In Section~\ref{section:overview} we provide an overview of the proof of exponential stability (Theorem~\ref{theorem:stability-informal}), and how this leads to a quasipolynomial-time planning algorithm (Theorem~\ref{thm:main-intro}). In Section~\ref{section:stability}, we formally prove Theorem~\ref{theorem:stability-informal}. In Section~\ref{section:planning}, we describe the quasipolynomial-time planning algorithm and prove Theorem~\ref{thm:main-intro}. Finally, in Section~\ref{section:lower-bounds}, we complement our results with various lower bounds: that our algorithm is optimal under the Exponential Time Hypothesis (Theorem~\ref{theorem:lb-intro}), that weak observability is not a strong enough assumption (Proposition~\ref{prop:weak-observability-lower-bound}), and that our belief contraction bound is optimal in $\gamma$.

\section{Preliminaries}\label{section:preliminaries}
In this section we introduce our notation and present relevant background. For a vector $v \in \BR^d$ (for some $d \in \BN$), we denote its components by $v(1), v(2), \ldots, v(d)$. For a sequence $(v_1,\dots,v_H)$, we let $v_{a:b}$ refer to the subsequence $(v_a,v_{a+1},\dots,v_b)$. Throughout the paper $\log$ refers to the natural logarithm. For any discrete set $S$, we let $\Delta^S$ refer to the subset of $\RR^S$ consisting of distributions on $S$. For discrete distributions $P,Q$, we write $P \ll Q$ if $Q(x)=0$ implies that $P(x)=0$. 

\subsection{POMDPs}
We consider the setting of finite-horizon partially observable Markov decision processes (POMDPs). A POMDP is specified by a tuple $(H, \MS, \MA, \MO, b_1, R, \BT, \BO)$. Here $H \in \BN$ gives the horizon length, $\MS$ is a finite set of \emph{states} of size $S := |\MS|$, $\MA$ is a finite set of \emph{actions} of size $A := |\MA|$, $\MO$ is a finite set of observations of size $O := |\MO|$, and $b_1 \in \Delta^\MS$ is the initial distribution over states. It remains to describe the transition kernels $\BT$, the observation matrices $\BO$, and the reward functions $R$: for each $h \in [H-1]$, $\BT_h(\cdot | x, a) \in \Delta^\MS$ is the distribution of the next state after being at state $x \in \MS$ and taking action $a \in \MA$ at step $h$. For each $h \in \{2,\dots,H\}$, $\BO_h(\cdot | x) \in \Delta^\MO$ is the distribution of the observation produced when at state $x \in \MS$ and step $h \in [H]$. Finally, for each $h \in \{2,\dots,H\}$, $R_h : \MO \ra [0,1]$ is the \emph{reward function} that gives the reward received after seeing an observation at step $h$. With slight abuse of notation, we will write $\BT_h(a) \in \BR^{\MS \times \MS}$ to denote the matrix given by $\BT_h(a)_{x,x'} = \BT_h(x | x',a)$, and $\BO_h \in \BR^{\MS \times \MO}$ to denote the matrix given by $(\BO_h)_{x,y} = \BO_h(y|x)$. Further, for a distribution $b \in \Delta^\MS$ and $h \in \{2,\dots,H\}$, we will write $\BO_h(y | b) = \sum_{x \in \MS} b(x) \cdot \BO_h(y|x) = (\BO_h^\t b)(y)$. 

We consider the \emph{planning} problem in POMDPs, meaning that the data of the POMDP, namely $(H, \MS, \MA, \MO, b_1, R, \BT, \BO)$ are all known to the algorithm. The algorithm interacts with the POMDP in the following fashion: at the beginning of the interaction, a state $x \sim b_1$ is drawn from the initial state distribution. At each step $1 \leq h <H$, the algorithm takes action $a_h \in \MA$, the POMDP transitions to a new state $x_{h+1} \sim \BT_h(\cdot | x_h, a_h)$, the algorithm sees the observation $o_{h+1} \sim \BO_{h+1}(x_{h+1})$, and receives reward $R_{h+1}(o_{h+1})$. Crucially, the algorithm does not observe the underlying states $x_1, x_2, \ldots, x_H$.

Let $\CH_h := (\MA \times \MO)^{h-1}$ denote the set of \emph{histories} at step $h$, namely the set of action/observation sequences $(a_1, o_2, a_2,\dots, a_{h-1}, o_h)$ (written for conciseness as $(a_{1:h-1},o_{2:h})$) the agent has access to before choosing the action $a_h$. The goal of the planning algorithm is to output a \emph{policy} for the agent, namely a collection $(\pi_h)_{h \in [H-1]}$ of functions $$\pi_h: \CH_h \to \MA.$$
The \emph{value} of policy $\pi$, denoted $V^\pi$, is the expected total reward received by an agent interacting with the POMDP as described above, and using $\pi_h$ to choose the actions $a_h \in \MA$. Formally, we have $V^\pi = \E_{(o_2,o_3, \ldots, o_H) \sim \pi} \left[ \sum_{h=2}^H R_h(o_h)\right]$.  In general, the policy could be randomized, but as is the case for (fully observed) Markov decision processes, the value of a POMDP can always be maximized by a deterministic policy. Unlike for fully-observed MDPs, however, the optimal policy in a POMDP typically takes actions depending on not just the most recent observation but in fact the entire history. While in general, describing the optimal policy can thus take exponential space, it will turn out that for observable POMDPs, a near-optimal policy can be described in only quasipolynomial space.

\subsection{Belief states, Policies, and Value functions}
Next we make some basic definitions regarding the updates of \emph{belief states} in POMDPs. Informally, the belief state at some step $h$ is the conditional distribution on states at that step, given the sequence of actions and observations that we have observed up to step $h$. The update to the belief state in response to each new action and observation pair is formalized in the below definition.
\begin{definition}[Belief state update]
  \label{def:bel-update}
For each $h \in \{2,\dots,H\}$, the \emph{Bayes operator} is $B_h: \Delta^\MS \times \MO \to \Delta^\MS$ defined by $$B_h(b; y)(x) = \frac{\bO_h(y|x)b(x)}{\sum_{z \in \MS} \bO_h(y|z)b(z)}.$$
For each $h \in [H-1]$, the \emph{belief update operator} $U_h: \Delta^\MS \times \MA \times \MO \to \Delta^\MS$, is defined by $$U_h(b;a,y) =  B_{h+1}(\bT_h(a) \cdot b;y).$$ In words, $U_h(b;a,y)$ describes the belief update we make when starting at belief state $b$, taking action $a \in \MA$, and then observing $y \in \MO$. Explicitly, we have, for $x \in \MS$,
\begin{align}
U_h(b;a,y)(x) = \frac{\BO_{h+1}(y|x) \cdot \sum_{x' \in \MS} b(x') \cdot \BT_h(x | x', a)}{\sum_{z \in \MS} \BO_{h+1}(y|z) \sum_{x' \in \MS} b(x') \cdot \BT_h(z | x',a)}\nonumber.
\end{align}
\end{definition}

We use the notation $\belief_h$ to denote belief states: the belief state at step $h = 1$ is defined as $\belief_1(\emptyset) = b_1$, and for any $2 \leq h \leq H$ and any action/observation sequence $(a_{1:h-1},o_{2:h})$, we inductively define the belief state
\begin{align}
  \label{eq:belief-defn}
  \belief_h(a_{1:h-1},o_{2:h}) = U_{h-1}(\belief_{h-1}(a_{1:h-2}, o_{2:h-1}); a_{h-1}, o_h).
  \end{align}
This definition ensures that for any policy, the distribution of $x_h$ conditioned on the history $(a_{1:h-1},o_{2:h})$ and the initial distribution $x_1 \sim b_1$, is precisely $\belief_h(a_{1:h-1},o_{2:h})$. Thus, the value of policy $\pi$ can be defined in terms of belief states.
\begin{definition}[Value function of a policy]
  \label{def:pol-valfn}
For any policy $\pi$, and $h \in [H]$, inductively define the value function $V^\pi_h: \CH_h \to \RR$ at time $h$ by $$V^\pi_h(a_{1:h-1},o_{2:h}) = \EE_{y \sim (\bO_{h+1})^\t \bT_h(a_h)\belief_h(a_{1:h-1},o_{2:h})} R(y) + V^\pi_{h+1}((a_{1:h-1},a_h), (o_{2:h},y))$$ where $a_h = \pi_h(a_{1:h-1},o_{2:h})$ and $V^\pi_H = 0$. It can be seen that $V^\pi_1(\emptyset)$ is precisely $V^\pi$, the value of $\pi$.
\end{definition}

Moreover, the optimal policy can be described as follows. 

\begin{proposition}[Bellman optimality]\label{prop:optimal-valfn}
Let $V^*_H(a_{1:H-1}, o_{2:H}) = 0$ for all $a,o$. For any $1 \leq h < H$, define
$$Q^*_h((a_{1:h-1},a), o_{2:h}) = \EE_{y \sim (\bO_{h+1})^\t \bT_h(a) \belief_h(a_{1:h-1},o_{2:h})} R(y) + V^*_{h+1}((a_{1:h-1},a),(o_{2:h},y)).$$
$$\pi^*_h(a_{1:h-1},o_{2:h}) = \argmax_{a \in \MA} Q^*_h((a_{1:h-1},a), o_{2:h}).$$
$$V^*_h(a_{1:h-1}, o_{2:h}) = \max_{a \in \MA} Q^*_h((a_{1:h-1},a), o_{2:h}).$$
Then $\pi^*$ is an optimal policy.
\end{proposition}

We also define approximate belief states. For $1 \leq h \leq H$, we define $$\sbelief_h(\emptyset) = \begin{cases} b_1 & \text{ if } h = 1 \\ \Unif(\MS) & \text{ otherwise} \end{cases}.$$
Then for any $1 \leq h-t < h \leq H$ and any action/observation sequence $(a_{h-t:h-1}, o_{h-t+1:h})$, we inductively define \begin{align}\label{eq:approximate-belief-defn}\sbelief_{h}(a_{h-t:h-1},o_{h-t+1:h}) = U_{h-1}(\sbelief_{h-1}(a_{h-t:h-2},o_{h-t+1:h-1}); a_{h-1}, o_h).\end{align}

\subsection{Divergences}
In this section we recall several divergence measures between distributions. 
Recall that we let $\log$ refer to the natural logarithm.

\begin{definition}
Define $f_\text{KL}: (0,\infty) \to \RR$ by $f(x) = x - \log(x) - 1$.
\end{definition}

\begin{definition}
For discrete distributions $P,Q$ with $P \ll Q$ (i.e. $P[Q=0] = 0$), the KL-divergence is $$\kld{P}{Q} := \EE_{x \sim P} \left[f_\text{KL}\left(\frac{Q(x)}{P(x)}\right) \right]= \sum_x P(x) \log \frac{P(x)}{Q(x)}.$$ 
The total variation distance is $$\TV(P,Q) := \frac{1}{2}\norm{P - Q}_1 = \frac{1}{2} \EE_{x \sim P} \left[ \left|\frac{Q(x)}{P(x)} - 1\right|\right] = \frac{1}{2}\sum_x |P(x) - Q(x)|.$$
The $\chi^2$-divergence is $$\chisq{P}{Q} := \EE_{x \sim Q} \left[\left(\frac{P(x)}{Q(x)}\right)^2 - 1\right] = \E_{x \sim P} \left[ \frac{P(x)}{Q(x)} - 1 \right].$$
The $2$-\Renyi divergence is $$\ren{P}{Q} := \log \EE_{x \sim P} \left[\frac{P(x)}{Q(x)}\right] = \log \EE_{x \sim Q} \left[\left(\frac{P(x)}{Q(x)}\right)^2 \right]= \log(1 + \chisq{P}{Q}).$$
The squared Hellinger distance $H^2(P,Q)$ is defined by $$H^2(P,Q) = 1 - \EE_{x \sim P} \sqrt{\frac{Q(x)}{P(x)}}.$$
All of the above are examples of $f$-divergences. For a convex function $f: [0,\infty) \to \RR$ with $f(1) = 0$, and any discrete distributions $P,Q$ with $P \ll Q$, the $f$-divergence is defined as $$D_f(P||Q) = \EE_{x \sim Q} f\left(\frac{Q(x)}{P(x)}\right).$$
\end{definition}

We will often make use of the following two inequalities.

\begin{lemma}[Pinsker's inequality]
For distributions $P,Q$, $$2\TV(P,Q)^2 \leq \kld{P}{Q}.$$
\end{lemma}

\begin{lemma}[Data processing inequality; \cite{polyanskiy2014lecture}]
For any $f$-divergence, for distributions $P,Q$ and a random transformation $K$, it holds that $$D_f(K \circ P||K \circ Q) \leq D_f(P || Q).$$
\end{lemma}

\section{Technical overview}\label{section:overview}

In this section, we outline our quasipolynomial planning algorithm and the proof of its correctness. Planning in POMDPs is in general computationally hard because of the \emph{curse of history}: the optimal policy may depend on the entire history, and therefore may have description complexity exponential in the horizon length $H$. In observable POMDPs, this is still the case, but we show that there is always a \emph{nearly-optimal} policy which chooses an action based on only a short, logarithmic-length window of recent observations and actions (and which can be computed in quasi-polynomial time).

\paragraph{Discretization of belief space.} Any sequence of actions and observations $(a_{1:h-1},o_{2:h})$, together with an initial state distribution $b_1$, induces a belief state $\belief_h(a_{1:h-1}, o_{2:h})$ (defined formally in (\ref{eq:belief-defn})), which is the posterior distribution on the latent state at time $h$, given the history $(a_{1:h-1}, o_{2:h})$. There is always an optimal policy which is a function of this belief state, rather than a function of the whole history. Of course, since a belief state is a vector in the continuous and high-dimensional space $\Delta^S$, this fact does not yield an efficient planning algorithm. The classical approach of discretizing the belief space and maintaining a value function only on a total variation net of $\Delta^S$ achieves $\epsilon$-suboptimality with time complexity $(\poly(H)/\epsilon)^S$. Alternatively, computing the value function on all histories by Bellman updates finds the optimal policy with time complexity $(OA)^H$; this can be thought of as a (lossless) discretization of the belief space of cardinality $(OA)^H$. Unfortunately, both approaches require exponential time. Moreover, even under $\gamma$-observability, there may not be a sub-exponential sized net on $\Delta^S$ such that every belief state reachable by a sequence of (positive-probability) observations and actions lies within total variation distance $\epsilon$ of some belief in the net (Example~\ref{ex:large-net-needed}). However, we show that under observability, there is a quasipolynomial-size discretization of the belief space $\Delta^S$ which is close to reachable beliefs \emph{on average} (under any policy), and which as a result can be used to compute an $\epsilon$-suboptimal policy.

\paragraph{Short memory.} For any window length $t < h$, we can consider trying to approximate $b_h := \belief_h(a_{1:h-1}, o_{2:h})$ using only the information present in the last $t$ observations and actions. Concretely, for any prior $b'_{h-t}$ on the state at time $h-t$, the history suffix $(a_{h-t:h-1},o_{h-t+1:h})$ induces a posterior distribution $b'_h = b_h'(a_{h-t:h-1}, o_{h-t+1:h})$ on the state at time $h$. Our main technical result is that under observability, if $b'_{h-t}$ is the \emph{uniform} prior on states (meaning that $b_h'$ is exactly the \emph{approximate belief state} $\sbelief_h(a_{h-t:h-1}, o_{h-t+1:h})$ defined in (\ref{eq:approximate-belief-defn})), then $b_h$ (the true belief state) and $b'_h$ (the approximate belief state) are close in expectation over histories. Quantitatively, we show (Theorem \ref{theorem:stability-informal}) that if $t \geq \widetilde{\Omega}(\gamma^{-4} \log(1/\epsilon))$ (hiding factors logarithmic in the instance size) then $\EE \tvd{b_h}{b'_h} \leq \epsilon$.

This approximation result suggests a concise discretization of belief space: the set of $\epsilon/\poly(H)$-approximate belief states $b_h'$ defined above (indexed by all possible choices of $a_{h-t:h-1}, o_{h-t:1:h}$), 
which has cardinality exponential in $t = \widetilde{\Omega}(\log(H/\epsilon))$ and therefore quasipolynomial in the instance size. One approach is to explicitly compute these belief states, and apply Approximate Point-Based Value Iteration \cite{pineau2006anytime}. 
However, it is simpler to just represent each belief state in the discretization by the length-$t$ history that induces it. 

That is, instead of computing the optimal value function $V^*(a_{1:h-1}, o_{2:h})$, which is described by the following Bellman update: \begin{equation} V^*_h(a_{1:h-1}, o_{2:h}) = \max_{a_h \in \MA} \EE_{o_{h+1} \sim (\bO_{h+1})^\t \bT_h(a_h) b_h} \left[R(o_{h+1}) + V^*_{h+1}(a_{1:h}, o_{2:h+1})\right],\label{eq:bellman-optimal}\end{equation}
we compute an approximate value function $\hat{V}_h(a_{h-t:h-1},o_{h-t+1:h})$:
\begin{equation} \hat{V}_h(a_{h-t:h-1},o_{h-t+1:h}) = \max_{a_h \in \MA} \EE_{o_{h+1} \sim (\bO_{h+1})^\t \bT_h(a) b'_h} \left[R(o_{h+1}) + \hat{V}_{h+1}(a_{h-t+1:h},\dots,o_{h+t+2:h+1})\right],\label{eq:bellman-approximate}\end{equation}
where $b_h' = \sbelief_h(a_{h-t:h-1}, o_{h-t+1:h})$ is the approximate belief state. 
Note that in Equation~\ref{eq:bellman-approximate}, the expectation is with respect to this approximate belief state, which by definition depends only on the last $t$ actions and observations, so the update is well-defined (i.e. $\hat{V}_h$ is indeed a function of only $a_{h-t:h-1}$ and $o_{h-t+1:h}$, not the entire history). Moreover, by our approximation result, we can inductively show that $\hat{V}_h$ and $V^*_h$ are close in expectation over any policy, so in particular $\hat{V}_1(\emptyset)$ and $V^*_1(\emptyset)$ are close. Finally, $\hat{V}_h$ defines a policy (via the implicit $Q$-function), which we can inductively show has value close to $\hat{V}_h$ by a similar inductive argument. Overall, we are able to compute an $\epsilon \cdot \poly(H)$-suboptimal policy in time $(OA)^{O(\gamma^{-4}\log(S/\epsilon))}$, up to polynomial factors. Equivalently, we get an $\epsilon$-suboptimal policy in time $(OA)^{O(\gamma^{-4} \log(SH/\epsilon))}$.

\paragraph{Belief contraction.} It remains to prove the belief state approximation result of Theorem \ref{theorem:stability-informal}. Let $b_{h}$ be the true belief state at some arbitrary time $h$, given the history so far, and let $b'_{h}$ be any other distribution over states. After taking action $a_h$ and seeing observation $o_{h+1}$, denote the output of the belief update operator, $U_h(b_h; a_h, o_{h+1})$ (Definition \ref{def:bel-update}), by $b_{h+1}$; recall that $b_{h+1}$ is computed from $b_h$ by applying the transition matrix $\bT_h(a_h)$ and the Bayes update by $o_{h+1}$. The same operation can be applied to $b'_h$ to obtain a distribution $b'_{h+1} = U_h(b_h'; a_h, o_{h+1})$, which is the posterior distribution on the state at time $h+1$ given prior $b'_h$, action $a_h$, and observation $o_{h+1}$. We want to show that under observability, the beliefs tend to contract, i.e. $b'_{h+1}$ and $b_{h+1}$ tend to be closer than $b_h$ and $b'_h$.

Proving contraction requires bounding some measure of divergence between distributions. We present two different arguments, using two different divergences (KL-divergence and $2$-\Renyi divergence), which give quantitatively incomparable results. In this overview, we start by outlining the KL-divergence argument, which is simpler but contains the main ideas.

\paragraph{Contraction via KL.} Fix some step $h_0$ and window length $t > 0$, and (as in the setting of Theorem \ref{theorem:stability-informal}) suppose that $b'_{h_0-t}$ is uniform on states; then the initial KL-divergence $\kld{b_{h_0-t}}{b'_{h_0-t}}$ is bounded above by $\log(S)$. By the data processing inequality, applying the transition matrix to each of the true belief distribution $b_h$ and the approximate belief distribution $b_h'$, for any $h \geq h_0-t$, cannot increase the divergence between them. However, the Bayes update (see Definition \ref{def:bel-update}) is not a random transformation, and in fact it can increase the divergence for some observations (Example~\ref{ex:divergence-increase}). For prior distribution $b$ and observation $o_{h} = y$, recall that the Bayes update at step $h$ is denoted $B_{h}(b;y)$. Then the chain rule for KL-divergence implies the following bound on the \emph{expected} KL-divergence, for any two distributions $b,b' \in \Delta^S$ and any $h \in [H]$:
\begin{align*}
\EE_{y \sim \bO_{h}^\t b} \left[\kld{B_{h}(b;y)}{B_{h}(b';y)}\right]
&= \kld{b}{b'} - \kld{\bO_{h}^\t b}{\bO_{h}^\t b'}
\end{align*}
By observability and Pinsker's inequality, we get the upper bound 
\begin{align*}
\EE_{y \sim \bO_{h}^\t b} \left[\kld{B_h(b;y)}{B_h(b';y)}\right]
&\leq \kld{b}{b'} - \frac{1}{2}\norm{\bO_{h}^\t b- \bO_{h}^\t b'}_1^2 \\
&\leq \kld{b}{b'} - \frac{\gamma^2}{2} \norm{b - b'}_1^2
\end{align*}

Unfortunately, the reverse of Pinsker's inequality does not hold in general, so the decrement $\gamma^2 \norm{b - b'}_1^2/2$ cannot be lower bounded by $\Omega(\gamma^2 \kld{b}{b'})$. When $\norm{b/b'}_\infty$ is bounded, it can be shown that $\norm{b - b'}_1 \geq \Omega(\kld{b}{b'})$. However, this loses a factor of $\kld{b}{b'}$ compared to if Pinsker's inequality were tight; we get the recursion
\begin{equation}
\EE_{y \sim \bO_{h}^\t b} \left[\kld{B_h(b;y)}{B_h(b';y)}\right]
\leq \kld{b}{b'} - c\gamma^2 \kld{b}{b'}^2,
\label{eq:slow-contraction}
\end{equation}
which may be seen to imply a polynomial rate of contraction (rather than the desired exponential rate): $\EE \kld{b_{h_0}}{b'_{h_0}} \leq \epsilon$ for window length $t \geq \widetilde{\Omega}(1/(\gamma^2 \ep))$. 
This is essentially the result (and proof technique) obtained in \cite{even2007value}. However, this rate does not suffice for the planning algorithm, because we ultimately need an error bound of $\epsilon/\poly(H)$, in which case we would need the window length $t$ to be larger than the horizon length. Moreover, Equation~\ref{eq:slow-contraction} is tight (Example~\ref{ex:no-linear-rate}), so we cannot achieve an exponential rate of contraction by reasoning about the expected KL-divergence at each step.

Instead, we appeal to a variance-based argument. We show that for any distributions $b,b' \in \Delta^S$, either the expected absolute difference between $\kld{B_h(b;y)}{B_h(b';y)}$ and its expectation is lower bounded by $\Omega(\gamma \kld{b}{b'})$ (i.e., $\kld{B_h(b;y)}{B_h(b;y)}$ is fairly anti-concentrated), or else a stronger reverse Pinsker inequality holds, namely $\norm{b-b'}_1 \geq \Omega(1 \land \sqrt{\kld{b}{b'}})$. 
Both cases imply a decay in the expected square-root of the KL-divergence: \begin{equation} \EE \sqrt{\kld{B_h(b,y)}{B_h(b',y)}} \leq \left(1 - \Omega\left(\frac{\gamma^2}{1 \lor \kld{b}{b'}}\right)\right)\sqrt{\kld{b}{b'}}.\label{eq:kl-sqrt-contraction}\end{equation}
To upper bound $\kld{b_{h_0}}{b_{h_0}'}$, we apply
the contraction inequality (\ref{eq:kl-sqrt-contraction})  with $b = \BT_{h-1}(a_{h-1}) \cdot b_{h-1},\ b' = \BT_{h-1}(a_{h-1}) \cdot b_{h-1}'$ for all $h > h_0 - t$. By a martingale-based argument, $\kld{b_h}{b_h'}$ is upper bounded by $\widetilde{\Omega}(\log(1/\epsilon))$ at all steps $h$, except with probability at most $\epsilon$. As a result, this recursion implies that the divergence contracts to $\epsilon^2$ in $t = \widetilde{O}(\gamma^{-2} \cdot \log^2(1/\epsilon))$ steps (except with probability $\epsilon$). By a final application of Pinsker's inequality, and since TV-distance is always at most $1$, we get that the total variation distance after $t$ steps can be bounded in expectation by $O(\epsilon)$. 

\paragraph{Linear Rates via $2$-\Renyi.} The KL-divergence argument gives optimal dependence on $\gamma$ (Proposition~\ref{prop:belief-contraction-lower-bound}) but does not quite achieve a linear rate. Using KL-divergence as the measure of divergence, this seems hard to avoid. Under the stochastic recursion described in Equation~\ref{eq:kl-sqrt-contraction}, the KL-divergence essentially decreases by a constant additive factor when larger than $1$, and by a constant multiplicative factor when smaller than $1$ (more precisely, there is a related potential function of KL which has these properties in expectation). Intuitively, one would expect that since the divergence starts off at $\log(S)$, it then takes $O(\log S)$ steps to reach $1$, and then $O(\log(1/\epsilon))$ steps to reach $\epsilon^2$. Unfortunately, since the recursion is only in expectation, this intuition may not be true: one might imagine that after the divergence reaches $1$, it might ``jump" to a large value with small probability. Indeed, there is a random process $(X_n)_n$ which satisfies the described recursion, and satisfies the upper bound $\sup_n X_n \leq \widetilde{O}(\log(1/\epsilon))$, but such that the smallest $n$ so that $\EE \min(X_n,1) \leq \ep$ is $n \geq \widetilde{\Omega}(\log^2(1/\epsilon))$. 

To avoid the issue of large jumps, we would like to control higher moments of the divergence, which can be accomplished by bounding $\EE \exp(\kld{b_h}{b_h'})$, which could potentially decrease by a constant multiplicative factor each step $h$ even when $\kld{b_h}{b_h'}$ is large. Indeed, this is how we obtained the upper bound of $\widetilde{\Omega}(\log(1/\epsilon))$: the KL-divergence is upper bounded by the $\infty$-\Renyi divergence, and $\exp(D_\infty(b_h||b_h'))$ is a supermartingale. Unfortunately, both $\exp(\kld{b_h}{b_h'})$ and $\exp(D_\infty(b_h||b_h'))$ are hard to manipulate. This motivates the use of the $2$-\Renyi divergence, which has the useful property that $\exp(D_2(b||b')) - 1 = \chisq{b}{b'}$.

The proof of contraction via $D_2(b||b')$ is much more intricate than the proof using $\kld{b}{b'}$, and has more cases, but these cases use the same main ideas: either the divergence has large absolute deviation from its mean, which implies that there is some event under which the divergence decreases substantially, or else we obtain a ``reverse $f$-divergence inequality" lower bounding the total variation distance $\norm{b-b'}_1$, which together with observability again yields a decrease in the divergence. As with the KL-divergence proof, which used $\EE \sqrt{\kld{b_h}{b_h'}}$ as the potential, here we also need an appropriate potential. It turns out that the correct potential is $$\EE \sqrt{\exp(D_2(b_h||b_h')/4) - 1},$$ which we are ultimately able to show decays at a linear rate. It is bounded above by $S$ at step $h = h_0-t$ (since $b_{h_0-t}'$ is assumed to be uniform), and it upper bounds the total variation distance, so this decay suffices to prove that beliefs contract at a linear rate in total variation distance. Unfortunately, the dependence on $\gamma$ is sub-optimal ($\gamma^{-4}$ instead of $\gamma^{-2}$), which is why our two approaches are incomparable, but for constant $\gamma$ this approach achieves the optimal rate.

\section{Exponential stability of belief states}\label{section:stability}

In this section, we prove that belief states are exponentially stable in expectation, via two methods. We start with the comparatively simpler argument, based on KL-divergence, which shows that beliefs contract to (total variation) distance at most $\epsilon$ in $\widetilde{O}(\log^2(1/\epsilon))$ steps. We then present the complete argument, based on $2$-\Renyi divergence, which achieves a bound of $\widetilde{O}(\log(1/\epsilon))$ steps; this argument is more involved but uses essentially the same techniques. However, we note that the arguments are not directly comparable: the contraction result based on KL-divergence achieves a better dependence on the observability lower bound $\gamma$.

\subsection{KL divergence}

\begin{theorem}\label{theorem:stability-kl}
There is a constant $C \geq 1$ so that the following holds. Suppose that the POMDP satisfies Assumption~\ref{assumption:observability-intro} with parameter $\gamma$. Let $\epsilon > 0$. Fix a policy $\pi$ and indices $1 \leq h < h+t \leq H$. If $t \geq C\gamma^{-2}\log(S/\epsilon)\log(\log(S)/\epsilon)$, then $$\EE_{a_{1:h+t-1},o_{2:h+t} \sim \pi} \norm{\belief_{h+t}(a_{1:h+t-1},o_{2:h+t}) - \sbelief_{h+t}(a_{h:h+t-1}, o_{h+1:h+t})}_1 \leq \epsilon.$$
\end{theorem}

The key lemma in the proof of the above theorem is that beliefs contract under the Bayes operator, in expected square-root KL-divergence (specifically, in the sense of part (3) of the following lemma):

\begin{lemma}\label{lemma:bayes-contraction}
Let $b, b' \in \Delta^\MS$ with $b \ll b'$, and fix any $h \in \{2,\dots,H\}$. Then
\begin{enumerate}
\item (Weak contractivity) $\EE_{y \sim \bO^\t_h b} \kld{B_h(b;y)}{B_h(b';y)} \leq \kld{b}{b'}$
\item (Strict contractivity or variance) Either $$\EE_{y \sim \bO^\t_h b} \kld{B_h(b;y)}{B_h(b';y)} \leq \kld{b}{b'} - \frac{\gamma^2}{32}\min(\kld{b}{b'}, 1)$$
or there is some event $\ME \subset \MO$ such that $$\EE_{y \sim \bO^\t_h b} (\kld{B_h(b;y)}{B_h(b';y)} - \kld{b}{b'})\mathbbm{1}[y \in \ME] \leq -\frac{\gamma}{8} \kld{b}{b'}.$$
\item $\EE_{y \sim \bO^\t_h b_h} \sqrt{\kld{B_h(b;y)}{B_h(b';y)}} \leq \left(1 - \frac{\gamma^2}{2^{14}\max(1, \kld{b}{b'})}\right)\sqrt{\kld{b}{b'}}.$
\end{enumerate}
\end{lemma}

\begin{proof}
We start by showing that the main claim (3) follows from (1) and (2). Indeed, under the first case in claim (2), we have by Jensen's inequality that
\begin{align*}
\EE_{y \sim \bO^\t_h b} \sqrt{\kld{B_h(b;y)}{B_h(b';y)}}
&\leq \sqrt{\EE_{y \sim \bO^\t_h b} \kld{B_h(b;y)}{B_h(b';y)}} \\
&\leq \sqrt{\left(1 - \frac{\gamma^2}{32\max(1,\kld{b}{b'})}\right)\kld{b}{b'}} \\
&\leq \left(1 - \frac{\gamma^2}{64\max(1,\kld{b}{b'})}\right)\sqrt{\kld{b}{b'}},
\end{align*}
as desired. On the other hand, if the first case does not hold, then $$\mu := \EE_{y \sim \bO^\t_h b} \kld{B_h(b;y)}{B_h(b';y)} > (1 - \gamma^2/32)\kld{b}{b'}.$$
By the second case of claim (2) and claim (1),
\begin{align*}
\EE_{y \sim \bO^\t_h b} (\kld{B_h(b;y)}{B_h(b';y)} - \mu)\mathbbm{1}[y \in \ME]
&\leq -\frac{\gamma}{8}\kld{b}{b'} + \kld{b}{b'} - \mu \\
&\leq -\frac{\gamma}{16}\kld{b}{b'}.
\end{align*}
It then follows from Lemma~\ref{lemma:sqrt-contraction} and claim (1) that $$\EE_{y \sim \bO^\t_h b} \sqrt{\kld{B_h(b;y)}{B_h(b';y)}} \leq \left(1 - \frac{\gamma^2}{2^{14}}\right)\sqrt{\kld{b}{b'}}$$
as desired. It remains to prove claims (1) and (2).

\paragraph{Claim (1).} Let $P_X$ be the distribution $b$ over states, and let $Q_X$ be the distribution $b'$ over states. Let $P_{Y|X}$ be the kernel describing the random map from states to observations, i.e. $P_{Y|X=x} = \MO_h(\cdot|x)$. Let $P_{X,Y} = P_{Y|X}P_X$ and $Q_{X,Y} = P_{Y|X}Q_X$. Then $P_{X|Y=y} = B_h(b;y)$ and $Q_{X|Y=y} = B_h(b';y)$. Thus,
\begin{align}
\kld{b}{b'}
&= \kld{P_X}{Q_X} \nonumber \\
&= \kld{P_{X,Y}}{Q_{X,Y}} \tag{chain rule using $Q_{Y|X} = P_{Y|X}$} \\
&= \kld{P_Y}{Q_Y} + \EE_{y \sim P_Y} \kld{P_{X|Y=y}}{Q_{X|Y=y}} \tag{chain rule} \\
&\geq \EE_{y \sim P_Y} \kld{P_{X|Y=y}}{Q_{X|Y=y}} \tag{non-negativity of KL-divergence} \\
&= \EE_{y \sim \bO^\t_h b} \kld{B_h(b;y)}{B_h(b';y)}. \nonumber
\end{align}
\paragraph{Claim (2).} For each observation $y \in \MO$ define $$A(y) = \bO_h(y|b)(\kld{B_h(b;y)}{B_h(b';y)} - \kld{b}{b'}).$$
We need to show that either $$\sum_{y \in \MO} A(y) \leq -\frac{\gamma^2}{32}\min(\kld{b}{b'},1)$$ or $$\sum_{y \in \MO: A(y) \leq 0} A(y) \leq -\frac{\gamma}{8}\kld{b}{b'}.$$
Expanding $A(y)$, we can write
\begin{align}
A(y)
&= \bO_h(y|b) \sum_{x \in \MS} \frac{\bO_h(y|x)b(x)}{\bO_h(y|b)} \log \frac{\bO_h(y|x)b(x) / \bO_h(y|b)}{\bO_h(y|x)b'(x) / \bO_h(y|b')} - \bO_h(y|b) \sum_{x \in \MS} b(x) \log \frac{b(x)}{b'(x)} \nonumber\\
&= \sum_{x \in \MS} (\bO_h(y|x) - \bO_h(y|b))b(x) \log \frac{b(x)}{b'(x)} + \sum_{x \in \MS} \bO_h(y|x) b(x) \log \frac{\bO_h(y|b')}{\bO_h(y|b)} \nonumber\\
&= \sum_{x \in \MS} (\bO_h(y|x) - \bO_h(y|b))b(x) \log \frac{b(x)}{b'(x)} + \bO_h(y|b) \log \frac{\bO_h(y|b')}{\bO_h(y|b)} \label{eq:ay-expand}\\
&= B(y) + C(y),\nonumber
\end{align}
where $$B(y) = \sum_{x \in \MS} (\bO_h(y|x) - \bO_h(y|b))b(x) \log \frac{b(x)}{b'(x)} + \bO_h(y|b') - \bO_h(y\|b)$$
and $$C(y) = \bO_h(y|b) \log \frac{\bO_h(y|b')}{\bO_h(y|b)} + \bO_h(y|b) - \bO_h(y|b').$$
This decomposition was chosen so that $C(y) \leq 0$ for all $y$ (Lemma~\ref{lemma:c-nonpositivity}). As argued in Lemma~\ref{lemma:large-term-b}, it follows that if $\sum_{y \in \MO} |B(y)| \geq \frac{\gamma}{4} \kld{b}{b'}$, then $\sum_{y \in \MO: A(y) \leq 0} A(y) \leq -\frac{\gamma}{8}\kld{b}{b'}.$ Thus, we can assume that $\frac{\gamma}{4} \kld{b}{b'} \geq \norm{B}_1$. But we can write the vector $B$ as $$B = (\bO_h)^\t \sum_{x \in \MS} (\delta_x - b) b(x) \log \frac{b(x)}{b'(x)} + (\bO_h)^\t b' - (\bO_h)^\t b,$$
where $\delta_x \in \BR^\MS$ denotes the vector with $\delta_x(x') = 1$ if $x = x'$ and $\delta_x(x') = 0$ if $x \neq x'$. 
Thus, by Assumption~\ref{assumption:observability-intro}, it follows that \begin{align*}
\frac{1}{4}\kld{b}{b'}
&\geq \norm{\sum_{x \in \MS} (\delta_x - b) b(x) \log \frac{b(x)}{b'(x)} + b - b'}_1 \\
&= \sum_{z \in \MS} \left| b(z) \log \frac{b(z)}{b'(z)} - b(z) \kld{b}{b'} + b'(z) - b(z)\right| \\
&= \EE_{z \sim b} \left| f_\text{KL}\left(\frac{b'(z)}{b(z)}\right) - \kld{b}{b'}\right|
\end{align*}
By Lemma~\ref{lemma:reverse-pinsker}, it follows that $\norm{b-b'}_1 \geq \min(\frac{1}{4} \sqrt{\kld{b}{b'}}, \frac{1}{10})$. Thus, we get that
\begin{align*}
\sum_{y \in \MO} A(y)
&= -\kld{(\bO_h)^\t b}{(\bO_h)^\t b'} \tag{Chain rule; follows from (\ref{eq:ay-expand})} \\
&\leq -\frac{1}{2}\norm{(\bO_h)^\t b - (\bO_h)^\t b'}_1^2 \tag{Pinsker's inequality} \\
&\leq -\frac{\gamma^2}{2} \norm{b - b'}_1^2 \tag{Assumption~\ref{assumption:observability-intro}} \\
&\leq -\frac{\gamma^2}{32}\min(\kld{b}{b'}, 1)
\end{align*}
as desired.
\end{proof}

In the proof of the above lemma, we used the following two lemmas.

\begin{lemma}\label{lemma:c-nonpositivity}
For any $y \in \MO$, it holds that $C(y) \leq 0$.
\end{lemma}

\begin{proof}
We can write $$C(y) = - \bO_h(y|b) \cdot f\left(\frac{\bO_h(y|b')}{\bO_h(y|b)}\right)$$ which is at most $0$ by Lemma~\ref{lemma:f-function}.
\end{proof}

\begin{lemma}\label{lemma:large-term-b}
If $\norm{B}_1 := \sum_{y \in \MO} |B(y)| \geq \eta \kld{b}{b'}$, then $$\sum_{y \in \MO: A(y) \leq 0} A(y) \leq -\frac{\eta}{2} \kld{b}{b'}.$$
\end{lemma}

\begin{proof}
Let $T = \{y: B(y) \leq 0\}$. By assumption $$\sum_{y \in \bar{T}} B(y) - \sum_{y \in T} B(y) \geq \eta \kld{b}{b'},$$ but by inspection $$\sum_{y \in \MO} B(y) = 0.$$ Thus, $$\sum_{y \in T} B(y) \leq -\frac{\eta}{2} \kld{b}{b'}.$$ By Lemma~\ref{lemma:c-nonpositivity}, we have $A(y) \leq B(y)$ for all $y$, so the lemma follows.
\end{proof}

In particular, it immediately follows from (3) in Lemma~\ref{lemma:bayes-contraction} that the belief update operator is contractive.

\begin{corollary}\label{lemma:belief-update-contraction}
Let $b, b' \in \Delta^\MS$ with $b \ll b'$, and fix any $h \in [H-1]$. Then for any action $a \in \MA$, with expectation over $y \sim (\bO_{h+1})^\t \bT_h(a) \cdot b$, $$\EE \sqrt{\kld{U_h(b;a,y)}{U_h(b';a,y)}} \leq \left(1 - \frac{\gamma^2}{2^{14}\max(1,\kld{b}{b'})}\right)\sqrt{\kld{b}{b'}}.$$
\end{corollary}

\begin{proof}
Suppose $\kld{\bT_h(a) \cdot b}{\bT_h(a) \cdot b'} \leq 1$. Then by definition of the belief update operator $U_h$,
\begin{align}
\EE \sqrt{\kld{U_h(b;a,y)}{U_h(b';a,y)}}
&= \EE \sqrt{\kld{B_{h+1}(\bT_h(a) \cdot b; y)}{B_{h+1}(\bT_h(a) \cdot b'; y)}} \nonumber\\
&\leq \left(1 - \frac{\gamma^2}{2^{14}}\right)\sqrt{\kld{\bT_h(a)\cdot b}{\bT_h(a) \cdot b'}} \tag{Lemma~\ref{lemma:bayes-contraction}}\\
&\leq \left(1 - \frac{\gamma^2}{2^{14}}\right)\sqrt{\kld{b}{b'}}. \tag{data processing inequality}
\end{align}
Similarly, if $\kld{\bT_h(a) \cdot b}{\bT_h(a) \cdot b'} \geq 1$, then
\begin{align}
\EE \sqrt{\kld{U_h(b;a,y)}{U_h(b';a,y)}}
&= \EE \sqrt{\kld{B_{h+1}(\bT_h(a) \cdot b; y)}{B_{h+1}(\bT_h(a) \cdot b'; y)}} \nonumber\\
&\leq \sqrt{\kld{\bT_h(a) \cdot b}{\bT_h(a)\cdot b'}} - \frac{\gamma^2}{2^{14}\sqrt{\kld{\bT_h(a) \cdot b}{\bT_h(a)\cdot b'}}} \tag{Lemma~\ref{lemma:bayes-contraction}} \\
&\leq \sqrt{\kld{b}{b'}} - \frac{\gamma^2}{2^{14}\sqrt{\kld{b}{b'}}}. \tag{data processing inequality}
\end{align}
In either case, it follows that $$\EE \sqrt{\kld{U_h(b;a,y)}{U_h(b';a,y)}} \leq \left(1 - \frac{\gamma^2}{2^{14}\max(1,\kld{b}{b'})}\right)\sqrt{\kld{b}{b'}}$$ as claimed.

\end{proof}

To prove Theorem~\ref{theorem:stability-kl}, we need one more lemma, which states that the exponential of the $\infty$-\Renyi divergence is a supermartingale.

\begin{lemma}\label{lemma:infinity-ratio}
Let $b,b' \in \Delta^\MS$ with $b \ll b'$, and fix any $h \in [H-1]$. For any $a \in \MA$, $$\EE_{y \sim (\bO_{h+1})\bT_h(a) b} \norm{\frac{U_h(b;a,y)}{U_h(b';a,y)}}_\infty \leq \norm{\frac{b}{b'}}_\infty.$$
\end{lemma}

\begin{proof}
  Suppose that $\norm{b/b'}_\infty = M$. We compute
  \begin{align}
    &  \EE_{y \sim (\bO_{h+1})^\t \bT_h(a) b} \left[ \left\| \frac{U(b; a, y)}{U(b';a,y)} \right\|_\infty \right]\nonumber\\
    =& \sum_{y \in \MO} (\bO_{h+1})^\t_y \bT_h(a) b \cdot \max_{x \in \MS} \left\{ \frac{\frac{\bO_{h+1}(y|x) \cdot \sum_{x'} b(x') \cdot \bT_h(x|x',a)}{(\bO_{h+1})^\t_y \bT_h(a) b}}{\frac{\bO_{h+1}(y|x) \cdot \sum_{x'} b'(x') \cdot \bT_h(x|x',a)}{(\bO_{h+1})^\t_y \bT_h(a) b'}}\right\} \nonumber\\
    =& \sum_{y \in \MO} (\bO_{h+1})^\t_y \bT_h(a) b' \cdot \max_{x \in \MS} \left\{ \frac{\bO_{h+1}(y|x) \cdot \sum_{x'} b(x') \cdot \bT_h(x|x',a)}{\bO_{h+1}(y|x) \cdot \sum_{x'} b'(x') \cdot \bT_h(x|x',a)}\right\}\leq M \nonumber,
  \end{align}
  where the final inequality follows since for each $x \in \MS$,
  \begin{align}
\bO_{h+1}(y|x) \cdot \sum_{x'} b(x') \cdot \bT_h(x|x',a) \leq M \cdot \bO_{h+1}(y|x) \cdot \sum_{x'} b'(x') \cdot \bT_h(x|x',a)\nonumber
  \end{align}
  by assumption.
\end{proof}

\begin{proof}[\textbf{Proof of Theorem~\ref{theorem:stability-kl}}]
Fix some history $(a_{1:h-1},o_{2:h})$. We condition on this history throughout the proof. For $0 \leq n \leq t$ define the random variables $$b_{h+n} = \belief_{h+n}(a_{1:h+n-1},o_{2:h+n}),$$
$$b'_{h+n} = \sbelief_{h+n}(a_{h:h+n-1},o_{h+1:h+n}),$$
$$X_n = \kld{b_{h+n}}{b'_{h+n}}.$$
Then $$X_0 = \kld{\belief_h(a_{1:h-1},o_{2:h})}{\sbelief_h(\emptyset)} \leq \log(S)$$ since $\sbelief_h(\emptyset)$ is either uniform on $\MS$ (for $h>1$) or equal to $\belief_h(\emptyset)$ (for $h=1$). Moreover, let $M = \log(S/\epsilon)$. For any $0 \leq n < t$, we have that
\begin{align}
&\EE_{a_{h:h+n},o_{h+1:h+n+1} \sim \pi} \sqrt{X_{n+1}} \mathbbm{1}[X_0,\dots,X_{n+1} \leq M] \nonumber\\
&\leq \EE_{a_{h:h+n},o_{h+1:h+n+1} \sim \pi} \sqrt{X_{n+1}} \mathbbm{1}[X_0,\dots,X_n \leq M] \nonumber \\
&= \EE_{a_{h:h+n-1},o_{h+1:h+n} \sim \pi}
\Bigg[\mathbbm{1}[X_0,\dots,X_n\leq M] \EE_{o_{h+n+1} \sim (\bO_{h+n+1})^\t \bT_{h+n}(\pi_{h+n-1}(a_{1:h+n-1},o_{2:h+n})) b_{h+n}} \nonumber \\
&\sqrt{\kld{U_{h+n}(b_{h+n}; \pi_{h+n-1}(a_{1:h+n-1},o_{2:h+n}), o_{h+n+1})}{U_{h+n}(b'_{h+n}; \pi_{h+n-1}(a_{1:h+n-1},o_{2:h+n}), o_{h+n+1})}}\Bigg] \label{eq:split-expectation}\\
&\leq \EE_{a_{h:h+n-1},o_{h+1:h+n} \sim \pi} \Bigg[\mathbbm{1}[X_0,\dots,X_n\leq M] \left(1 - \frac{\gamma^2}{2^{14}\max(1,\kld{b_{h+n}}{b'_{h+n}})}\right)\sqrt{\kld{b_{h+n}}{b'_{h+n}}}\Bigg] \label{eq:apply-cor}\\
&\leq \left(1 - \frac{\gamma^2}{2^{14}M}\right) \EE_{a_{h:h+n-1},o_{h+1:h+n} \sim \pi} \left[ \mathbbm{1}[X_0,\dots,X_n\leq M] \sqrt{X_n}\right] \nonumber
\end{align}
where (\ref{eq:split-expectation}) uses that $b_{h+n}$ is the conditional distribution of $x_{h+n}$ given $a_{1:h+n-1},o_{2:h+n}$, and the recursive definition of the belief state functions $\belief_{h+n+1}$ and $\sbelief_{h+n+1}$; and (\ref{eq:apply-cor}) uses Corollary~\ref{lemma:belief-update-contraction}. By induction, we get $$\EE\left[\sqrt{X_t} \mathbbm{1}[X_1,\dots,X_t \leq M]\right] \leq \left(1 - \frac{\gamma^2}{2^{14}M}\right)^t\sqrt{\log(S)} \leq \epsilon$$
where the last inequality uses that $$t \geq \frac{2^{14}M}{\gamma^2} \log \frac{\sqrt{\log(S)}}{\epsilon}.$$
Now we consider the event that $\sup X_n > M$. We have that $\norm{b_{h+n}/b'_{h+n}}_\infty$ is a nonnegative supermartingale (Lemma~\ref{lemma:infinity-ratio}) with $\norm{b_h/b'_h}_\infty \leq S$, so by Ville's maximal inequality, $$\Pr\left(\sup_n \norm{b_{h+n}/b'_{h+n}}_\infty > S/\epsilon\right) \leq \frac{  \norm{b_{h}/b'_{h}}_\infty}{S/\epsilon} \leq \epsilon.$$ 
But it holds deterministically that $\kld{b_{h+n}}{b'_{h+n}} \leq \log \norm{b_{h+n}/b'_{h+n}}_\infty$ for each $n$, so $$\Pr(\sup_n X_n > M) \leq \epsilon.$$
Thus,
\begin{align}
\EE \norm{b_{h+t} - b'_{h+t}}_1
&= \EE \norm{b_{h+t} - b'_{h+t}}_1 \mathbbm{1}[\sup_n X_n \leq M] + \EE \norm{b_{h+t} - b'_{h+t}}_1 \mathbbm{1}[\sup_n X_n > M] \nonumber \\
&\leq \EE\left[\sqrt{2X_t} \mathbbm{1}[\sup_n X_n \leq M]\right] + \EE \norm{b_{h+t} - b'_{h+t}}_1 \mathbbm{1}[\sup_n X_n > M] \tag{Pinsker's inequality} \\
&\leq \epsilon\sqrt{2} + 2\Pr(\sup_n X_n > M) \nonumber\\
&\leq (2+\sqrt{2})\epsilon \nonumber.
\end{align}
Taking expectation over $(a_{1:h-1},o_{2:h})$ completes the proof of the theorem.
\end{proof}

\subsection{$2$-\Renyi divergence}

\begin{theorem}\label{theorem:stability-renyi}
There is a constant $C \geq 1$ so that the following holds. Suppose that the POMDP satisfies Assumption~\ref{assumption:observability-intro} with parameter $\gamma$. Let $\epsilon > 0$. Fix a policy $\pi$ and indices $1 \leq h < h+t \leq H$. If $t \geq C\gamma^{-4}\log(S/\epsilon)$, then $$\EE_{a_{1:h+t-1},o_{2:h+t} \sim \pi} \norm{\belief_{h+t}(a_{1:h+t-1},o_{2:h+t}) - \sbelief_{h+t}(a_{h:h+t-1}, o_{h+1:h+t})}_1 \leq \epsilon.$$
\end{theorem}

To prove Theorem \ref{theorem:stability-renyi}, instead of using KL divergence as an upper bound for the total variation distance between distributions $P,Q$ (as was done in the previous subsection), we use the quantity $\sqrt{\exp(\ren{P}{Q}/4-1)}$ based on the $2$-\Renyi divergence. The below lemma verifies that this quantity indeed upper bounds total variation distance (up to a constant factor). 
\begin{lemma}\label{lemma:d2-to-tv}
  Consider probability distributions $P,Q$. Then
  \begin{align}
\tvd{P}{Q} \leq 4 \cdot \sqrt{\exp(\ren{P}{Q}/4)-1}\nonumber.
  \end{align}
\end{lemma}
\begin{proof}
  We have
  \begin{align}
\exp(\ren{P}{Q}/4) - 1 \geq \frac{\ren{P}{Q}}{4} \geq \frac{\kld{P}{Q}}{4} \geq \frac{\tvd{P}{Q}^2}{8}\nonumber,
  \end{align}
  where the second inequality uses the monotonicity of R\'{e}nyi divergences and the final inequality uses Pinsker's inequality. The claimed result follows immediately.
\end{proof}

Lemma \ref{lemma:renyi-bayes-update-contraction} is the main piece in the proof of Theorem \ref{theorem:stability-renyi}; it proves contraction in expectation of the R\'{e}nyi potential at each time step.
\begin{lemma}\label{lemma:renyi-bayes-update-contraction}
  Suppose $b,b' \in \Delta^\MS$ with $b \ll b'$, and fix any $h \in \{2,\dots,H\}$. 
  Then 
  \begin{align}
\E_{y \sim \BO_h^\t b} \left[ \sqrt{\exp \left( \frac{\ren{B_h(b;y)}{B_h(b';y)}}{4} \right) - 1}\right] \leq \left( 1 - \gamma^4/2^{40} \right) \cdot \sqrt{\exp \left( \frac{\ren{b}{b'}}{4} \right) - 1}\nonumber
  \end{align}
\end{lemma}
\begin{proof}
  The lemma is proven by considering several cases regarding the value of $\ren{b}{b'}$. To begin, we make some general observations regarding the value of $\ren{B_h(b;y)}{B_h(b';y)}$, which will be helpful throughout the proof. 
  For each $y \in \MO$, define
  \begin{align}
A(y) =\ren{B_h(b;y)}{B_h(b';y)} \geq 0\nonumber.
  \end{align}
  Using the definition of $B_h(\cdot)$, we have
  \begin{align}
    A(y) =& \log \left(\E_{x \sim B_h(b;y)} \left[ \frac{B_h(b;y)(x)}{B_h(b';y)(x)} \right] \right)\nonumber\\
    = &  \log \left(\E_{x \sim B_h(b;y)} \left[ \frac{b(x) \cdot \BO_h(y|x)}{\BO_h(y|b)} \cdot \frac{\BO_h(y|b')}{b'(x) \cdot \BO_h(y|x)} \right] \right)\nonumber\\
    = & \log \left( \frac{\BO_h(y|b')}{\BO_h(y|b)} \right) + \log \E_{x \sim B_h(b;y)} \left[ \frac{b(x)}{b'(x)} \right]\label{eq:ay-decompose}.
  \end{align}
  Now define $G(y) := \E_{x \sim B_h(b;y)} \left[ \frac{b(x)}{b'(x)} \right] + \frac{\BO_h(y|b')}{\BO_h(y|b)} - 2$. Then by two applications of the inequality $\log(a) \leq a-1$ for $a>0$, we have
  \begin{align}
    A(y)    \leq & A(y) - \log \left( \frac{\BO_h(y|b')}{\BO_h(y|b)} \right) - 1 + \frac{\BO_h(y|b')}{\BO_h(y|b)} \nonumber \\
    = & \log \E_{x \sim B_h(b;y)} \left[ \frac{b(x)}{b'(x)} \right] - 1 + \frac{\BO_h(y|b')}{\BO_h(y|b)}\nonumber\\
    \leq & \E_{x \sim B_h(b;y)} \left[ \frac{b(x)}{b'(x)} \right] - 2 + \frac{\BO_h(y|b')}{\BO_h(y|b)} \nonumber \\
    =& G(y)\label{eq:ag-bound}.
  \end{align}
  Furthermore, we have 
  \begin{align}
    \E_{y \sim \BO_h^\t b} \left[ G(y) \right] =& -1 + \sum_{y \in \MO} \BO_h(y|b) \cdot \sum_{x \in \MS} \frac{b(x) \cdot \BO_h(y|x)}{\BO_h(y|b)} \cdot \frac{b(x)}{b'(x)} \nonumber\\
    =& -1 + \sum_{x \in \MS} \frac{b(x)^2}{b'(x)} \nonumber\\
    =& \exp(\ren{b}{b'}) - 1 = \chisq{b}{b'}\label{eq:gy-expectation-chisq}.
  \end{align}
  Also define $F(y) := \E_{x \sim B_h(b;y)} \left[ \frac{b(x)}{b'(x)} \right]$. Note that
  \begin{align}
    \E_{y \sim \BO_h^\t b} [F(y)] = \exp(\ren{b}{b'}) = \chisq{b}{b'} +1.\label{eq:exp-fy-chisq}
  \end{align}
  Moreover note that
  \begin{align}
    \E_{y \sim\BO_h^\t b} \left[ \exp(A(y)/2) \right] = & \E_{y \sim \BO_h^\t b} \left[ \sqrt{\frac{\BO_h(y|b')}{\BO_h(y|b)}} \cdot \sqrt{F(y)}\right]\nonumber\\
    \leq & \sqrt{ \E_{y \sim \BO_h^\t b}\left[ \frac{\BO_h(y|b')}{\BO_h(y|b)} \right] \cdot \E_{y \sim \BO_h^\t b}[F(y)]}\tag{Cauchy-Schwarz}\\
    = & \sqrt{ \chisq{b}{b'}+1} = \exp \left( \frac{\ren{b}{b'}}{2} \right) \label{eq:exp-ay-2},
  \end{align}
  where the penultimate equality (\ref{eq:exp-ay-2}) uses (\ref{eq:exp-fy-chisq}). Define $c_0 = 1/256$. 
  We next consider several cases:

  \paragraph{Case 1.} $\chisq{b}{b'} \leq c_0\gamma$. Here, we consider the following sub-cases:
  \paragraph{Case 1a.\xlabel[1a]{case:1a}} Suppose that
    \begin{align}
      \E_{y \sim \BO_h^\t b} \left[ \exp(A(y)/2) - 1 \right] \leq (1-c_0 \gamma^2) \cdot (\exp(\ren{b}{b'}/2) - 1 )\label{eq:case-1a-define}.
    \end{align}
    Then by Jensen's inequality we have
    \begin{align}
\E_{y \sim \BO_h^\t b} \left[ \sqrt{\exp(A(y)/2) -1} \right] \leq (1-c_0\gamma^2/2) \cdot \sqrt{\exp(\ren{b}{b'}/2) - 1}\nonumber.
    \end{align}
    Consider the function $f(x) := \left( \sqrt{x^2+1}-1 \right)^{1/2}$. By Lemma \ref{lem:24-alpha-cor}, we have
    \begin{align}
      \E_{y \sim \BO_h^\t b} \left[ \sqrt{\exp(A(y)/4)-1} \right] =& \E_{y \sim \BO_h^\t b} \left[ f\left(\sqrt{\exp(A(y)/2)-1}\right) \right] \nonumber\\
      \leq & (1-c_0\gamma^2/16) \cdot \sqrt{\exp(\ren{b}{b'}/4)-1}\nonumber,
    \end{align}
as desired.
  \paragraph{Case 1b.\xlabel[1b]{case:1b}} Suppose that the previous case does not hold, and that moreover there is some event $\ME \subset \MO$ so that
    \begin{align}
\E_{y \sim \BO_h^\t b} \left[ \One[y \in \ME] \cdot (G(y) - \chisq{b}{b'}) \right] \leq - 3c_0 \gamma \cdot \chisq{b}{b'}\label{eq:smallchisq-gamma}.
    \end{align}
    Set
    \begin{align}
      \mu := \E_{y \sim \BO_h^\t} \left[ \exp(A(y)/2) - 1 \right] \geq & (1 -c_0\gamma^2) \cdot (\exp(\ren{b}{b'}/2) - 1)\label{eq:not-case-1a}\\
      \geq & (1 - c_0 \gamma^2) \cdot \frac{\ren{b}{b'}}{2} \nonumber\\
      \geq & (1 - c_0 \gamma^2) \cdot \frac{1}{2} \cdot \left( \chisq{b}{b'}- \ren{b}{b'}^2 \right)\label{eq:mu-d2-chisq-bound}\\
      \geq & (1 - c_0 \gamma) \cdot \frac{1}{2} \cdot \left( \chisq{b}{b'} \right) - \chisq{b}{b'}^2/2\label{eq:mu-bound},
    \end{align}
    where (\ref{eq:not-case-1a}) uses that (\ref{eq:case-1a-define}) does not hold, (\ref{eq:mu-d2-chisq-bound}) uses that $\exp(a) -1 \leq a + a^2$ for $a \leq 1$ (with $a = \ren{b}{b'}$, using that $\ren{b}{b'} \leq \chisq{b}{b'} \leq c_0\gamma \leq 1$) and (\ref{eq:mu-bound}) uses that $\ren{b}{b'} \leq \chisq{b}{b'}$.  
    It is without loss of generality to assume that $A(y) \leq G(y) \leq \chisq{b}{b'}$ for all $y \in \ME$ (otherwise, by removing any $y$ satisfying $G(y) > \chisq{b}{b'}$ from $\ME$, we can decrease the left-hand side of (\ref{eq:smallchisq-gamma})). Thus, 
    \begin{align}
      & \E_{y \sim \BO_h^\t b} \left[ \One[y \in \ME] \cdot ((\exp(A(y)/2) - 1) - \mu) \right]\nonumber\\
    \leq  & \E_{y \sim \BO_h^\t b} \left[ \One[y \in \ME] \cdot ((\exp(A(y)/2) - 1) - \chisq{b}{b'}/2) \right] + c_0 \gamma/2 \cdot \chisq{b}{b'} + \chisq{b}{b'}^2/2 \label{eq:use-mu-bound}\\
      \leq & \E_{y \sim \BO_h^\t b} \left[ \One[y \in \ME] \cdot (A(y)/2 + A(y)^2/4 - \chisq{b}{b'}/2) \right] + c_0\gamma/2 \cdot \chisq{b}{b'} + \chisq{b}{b'}^2/2 \label{eq:ay-exp-ub}\\
      \leq & \E_{y \sim \BO_h^\t b} \left[ \One[y \in \ME] \cdot (G(y)/2 - \chisq{b}{b'}/2 + \chisq{b}{b'}^2/4 ) \right] + c_0\gamma/2 \cdot \chisq{b}{b'} + \chisq{b}{b'}^2/2\label{eq:use-ay-ub}\\
      \leq & -3c_0 \gamma \cdot \chisq{b}{b'} + c_0\gamma/2 \cdot \chisq{b}{b'} +  \frac{3}{4} \cdot \chisq{b}{b'}^2\label{eq:use-smallchisq-gamma}\\
      \leq & -3c_0 \gamma \cdot \chisq{b}{b'} + c_0\gamma/2 \cdot \chisq{b}{b'} + \frac{3}{4} \cdot c_0 \gamma \cdot \chisq{b}{b'} \leq  -c_0 \gamma \cdot \chisq{b}{b'}\label{eq:chisq-sq},
    \end{align}
    where:
    \begin{itemize}
    \item (\ref{eq:use-mu-bound}) uses (\ref{eq:mu-bound});
    \item (\ref{eq:ay-exp-ub}) uses the fact that $\exp(A(y)/2) - 1 \leq A(y)/2 + A(y)^2/4$ for $y \in \ME$ (since $A(y) \leq \chisq{b}{b'} \leq c_0 \gamma$ for such $y$);
    \item (\ref{eq:use-ay-ub}) uses (\ref{eq:ag-bound}), as well as the fact that $0 \leq A(y) \leq \chisq{b}{b'}$ for $y \in \ME$;
    \item (\ref{eq:use-smallchisq-gamma}) follows from (\ref{eq:smallchisq-gamma});
    \item  (\ref{eq:chisq-sq}) follows from the assumption that $\chisq{b}{b'} \leq c_0\gamma$.
    \end{itemize}

    Thus, by Lemma \ref{lemma:sqrt-contraction},and using that $\chisq{b}{b'} \geq \mu$,
    we obtain
    \begin{align}
      \E_{y \sim \BO_h^\t b} \left[ \sqrt{\exp(A(y)/2) - 1} \right] \leq & (1 - c_0^2 \gamma^2/32) \cdot \sqrt{\E_{y \sim \BO_h^\t b}[\exp(A(y)/2) - 1]}\nonumber\\
      \leq & (1 - c_0^2 \gamma^2/32) \cdot \sqrt{\exp(\ren{b}{b'}/2) - 1} \label{eq:use-ay2-ub},
    \end{align}
    where (\ref{eq:use-ay2-ub}) follows from (\ref{eq:exp-ay-2}). As in Case \xref{case:1a}, consider the function $f(x) := \left( \sqrt{x^2 + 1} - 1\right)^{1/2}$. 
    By Lemma \ref{lem:24-alpha-cor}, 
    \begin{align}
      \E_{y \sim \BO_h^\t b} \left[ \sqrt{ \exp \left( \frac{A(y)}{4} \right) - 1} \right] = &  \E_{y \sim \BO_h^\t b} \left[ f \left( \sqrt{\exp(A(y)/2) - 1} \right) \right]\nonumber\\
      \leq & (1-c_0^2 \gamma^2/256) \cdot f\left( \sqrt{\exp(\ren{b}{b'}/2)-1} \right)\nonumber\\
      = & (1-c_0^2 \gamma^2/256) \cdot\sqrt{ \exp(\ren{b}{b'}/4) - 1}\nonumber,
    \end{align}
as desired (note that $c_0^2/256 = 1/2^{24}$).
  \paragraph{Case 1c.} If both Cases \xref{case:1a} and \xref{case:1b} do not hold, then in particular (\ref{eq:smallchisq-gamma}) does not hold for any $\ME \subset \MO$, and thus, using (\ref{eq:gy-expectation-chisq}), it holds that
    \begin{equation}
\E_{y \sim \BO_h^\t b} \left[ |G(y) - \chisq{b}{b'}| \right] \leq 6c_0 \gamma \cdot \chisq{b}{b'}\label{eq:1c-condition}.
    \end{equation}
    Define
    \begin{align}
      G'(y) :=& \BO_h(y|b) \cdot (G(y) - \chisq{b}{b'})\nonumber\\
      =& \BO_h(y|b) \cdot \left( \sum_{x \in \MS} \frac{b(x)^2 \cdot \BO_h(y|x)}{\BO_h(y|b) \cdot b'(x)} -\sum_{x \in \MS} \frac{b(x)^2}{b'(x)} + \frac{\BO_h(y|b')}{\BO_h(y|b)} - 1 \right)\nonumber\\
      =& \sum_{x \in \MS} \BO_h(y|x) \cdot \frac{b(x)^2}{b'(x)} - \sum_{x \in \MS} \BO_h(y|b) \cdot \frac{b(x)^2}{b'(x)} + \BO_h(y|b') - \BO_h(y|b)\nonumber.
    \end{align}
    Thus, the vector $G' \in \BR^\MO$ may be written as
    \begin{align}
G' = \BO_h^\t \sum_{x \in \MS} (\delta_x - b) \cdot \frac{b(x)^2}{b'(x)} + \BO_h^\t b' - \BO_h^\t b\nonumber,
    \end{align}
    where $\delta_x \in \BR^\MS$ is the indicator vector of $x$. Thus, by Assumption \ref{assumption:observability-intro},
    \begin{align}
\| G' \|_1 \geq \gamma \cdot \left\| \sum_{x \in \MS} (\delta_x - b) \cdot \frac{b(x)^2}{b'(x)} + b' - b \right\|_1 = \gamma \cdot \sum_{z \in \MS} \left| b(z) \cdot \left( \frac{b(z)}{b'(z)} + \frac{b'(z)}{b(z)} - 2 - \chisq{b}{b'}\right) \right|\nonumber.
    \end{align}
    Note that $\| G'\|_1 = \E_{y \sim \BO_h^\t b} \left[ |G(y) - \chisq{b}{b'}| \right]$. Thus, by (\ref{eq:1c-condition}), and rewriting the above sum as an expectation over $b$, we have
    \begin{align}
6c_0 \cdot \chisq{b}{b'} \geq \E_{z \sim b} \left| \frac{b(z)}{b'(z)} + \frac{b'(z)}{b(z)} -2 - \chisq{b}{b'} \right|\nonumber.
    \end{align}
    By Lemma \ref{lem:h-chisq}, since $c_0 \leq 1/24$ and $\chisq{b}{b'} \leq c_0 \gamma \leq 1/6$, it holds that $\tvd{b}{b'} \geq \frac{1}{4\sqrt 2} \cdot \sqrt{\chisq{b}{b'}}$. Then
    \begin{align}
      \E_{y \sim \BO_h^\t b} \left[ A(y)\right] =& \E_{y \sim \BO_h^\t b} [\log F(y)] - \E_{y \sim \BO_h^\t b} \left[ \log \left( \frac{\BO_h(y|b)}{\BO_h(y|b')} \right)\right]\nonumber\\
      \leq & \log \E_{y \sim \BO_h^\t b} [F(y)] - \kld{\BO_h^\t b}{\BO_h^\t b'}\nonumber\\
      \leq & \ren{b}{b'} - \frac 12 \cdot \tvd{\BO_h^\t b}{\BO_h^\t b'}^2 \tag{Pinsker's inequality}\\
      \leq & \ren{b}{b'} - \frac{\gamma^2}{2} \cdot \tvd{b}{b'}^2 \tag{Assumption \ref{assumption:observability-intro}} \\
      \leq & \ren{b}{b'} - \frac{\gamma^2}{64} \cdot \chisq{b}{b'}\label{eq:ay-d2-gammachisq}.
    \end{align}
    As in Case \xref{case:1b}, set
    \begin{align}
\mu := \E_{y \sim \BO_h^\t b} [\exp(A(y)/2) - 1] \geq (1 - c_0 \gamma^2) \cdot (\exp(\ren{b}{b'}/2) - 1)\label{eq:mu-ay-lb},
    \end{align}
    where the inequality uses that (\ref{eq:case-1a-define}) does not hold. 
    Set $\ME := \{ y \in \MO: A(y) \leq \ren{b}{b'} \}$. Note that
    \begin{align}
      & \E_{y \sim \BO_h^\t b} \left[ \One[y \in \ME] \cdot (A(y) - \ren{b}{b'}) \right] \nonumber\\
      \leq & \E_{y \sim \BO_h^\t b} \left[ \One[y \in \ME] \cdot (A(y) - \ren{b}{b'}) \right] +  \E_{y \sim \BO_h^\t b} \left[ \One[y \in \bar\ME] \cdot (A(y) - \ren{b}{b'}) \right]\nonumber\\
      \leq & -\frac{\gamma^2}{64} \cdot\chisq{b}{b'} \label{eq:ay-d2-e},
    \end{align}
    where (\ref{eq:ay-d2-e}) uses (\ref{eq:ay-d2-gammachisq}). 
    Thus,
    \begin{align}
      & \E_{y \sim \BO_h^\t b} \left[ \One[y \in \ME] \cdot ((\exp(A(y)/2) - 1) - \mu) \right]\nonumber\\
      \leq & \E_{y \sim \BO_h^\t b} \left[ \One[y \in \ME] \cdot ((\exp(A(y)/2) - 1) - (\exp(\ren{b}{b'}/2) - 1) )\right] + c_0 \gamma^2 \cdot (\exp(\ren{b}{b'}/2) - 1)\label{eq:use-mu-ay-lb}\\
      \leq & \E_{y \sim \BO_h^\t b} \left[ \One[y\in \ME] \cdot (A(y)/2 - \ren{b}{b'}/2 ) \right] + c_0 \gamma^2 \cdot (\exp(\ren{b}{b'}/2) - 1)\label{eq:use-ab-exp-ineq}\\
      \leq & -\frac{\gamma^2}{128} \cdot \chisq{b}{b'}  + c_0 \gamma^2 \cdot (\exp(\ren{b}{b'}/2) - 1)\label{eq:chisq-gamma-bound}\\
      \leq & -\frac{\gamma^2}{256} \cdot (\exp(\ren{b}{b'}/2) - 1) \label{eq:ren-2-chisq} \\
      \leq & - \frac{\gamma^2}{256} \cdot \mu,\label{eq:ren-2-mu}
    \end{align}
    where (\ref{eq:use-mu-ay-lb}) uses (\ref{eq:mu-ay-lb}), (\ref{eq:use-ab-exp-ineq}) uses the fact that for real numbers $0 \leq a \leq b$, we have $\exp(a) - \exp(b) \leq a-b$,  (\ref{eq:chisq-gamma-bound}) uses (\ref{eq:ay-d2-e}), (\ref{eq:ren-2-chisq}) uses the fact that $\exp(\ren{b}{b'}/2) - 1 \leq \exp(\ren{b}{b'}) - 1 = \chisq{b}{b'}$ as well as that $c_0 \leq 1/256$, and (\ref{eq:ren-2-mu}) uses (\ref{eq:exp-ay-2}). Thus, by Lemma \ref{lemma:sqrt-contraction}, we have that
    \begin{align}
      \E_{y \sim \BO_h^\t b} \left[ \sqrt{\exp(A(y)/2) - 1} \right] \leq & \left( 1-  \frac{\gamma^4}{256^2 \cdot 32}\right)\cdot \sqrt{\mu}\nonumber\\
      \leq & \left( 1 - \frac{\gamma^4}{2^{21}} \right) \cdot \sqrt{\exp(\ren{b}{b'}/2) - 1}\nonumber.
    \end{align}
    As in Case \xref{case:1b}, for the (concave) function $f(x) = \left( \sqrt{x^2+1}-1\right)^{1/2}$, we have, using Jensen's inequality, the above bound, and Lemma \ref{lem:24-alpha}, that
    \begin{align}
      \E_{y \sim \BO_h^\t b} \left[ \sqrt{\exp (A(y)/4) -1} \right] \leq & f\left((1-\gamma^4/2^{21}) \cdot \sqrt{\exp(\ren{b}{b'}/2)-1} \right)\nonumber\\
      \leq & (1-\gamma^4/2^{24}) \cdot \sqrt{\exp(\ren{b}{b'}/4)-1}\nonumber,
    \end{align}
as desired.


  \paragraph{Case 2.} $\chisq{b}{b'} \geq c_0\gamma$. Here, using (\ref{eq:ay-decompose}) and the Cauchy-Schwarz inequality, we have
  \begin{align}
    \E_{y \sim \BO_h^\t b} \left[ \exp \left( \frac{A(y)}{4} \right) \right] =& \E_{y \sim \BO_h^\t b} \left[ \left( \frac{\BO_h(y|b')}{\BO_h(y|b)} \right)^{1/4} \cdot \left(F(y) \right)^{1/4} \right]\nonumber\\
    \leq & \sqrt{\E_{y \sim \BO_h^\t b}  \left[\left( \frac{\BO_h(y|b')}{\BO_h(y|b)} \right)^{1/2}\right] }\cdot\sqrt{ \E_{y \sim \BO_h^\t b} \left[ \sqrt{F(y)} \right]}\label{eq:exp-ay-4}.
  \end{align}

 Again, we consider the following sub-cases:

  \paragraph{Case 2a\xlabel[2a]{case:2a}.} Suppose there is some event $\ME \subset \MO$ so that
    \begin{align}
\E_{y \sim \BO_h^\t b} \left[ \One[y \in \ME] \cdot (F(y) - \exp(\ren{b}{b'})) \right] \leq -c_0 \gamma \cdot \chisq{b}{b'}\nonumber.
    \end{align}
    Then by Lemma \ref{lemma:sqrt-contraction} and (\ref{eq:exp-fy-chisq}),
    \begin{align}
      &      \E_{y \sim \BO_h^\t b} [\sqrt{F(y)}] \leq \left( 1 - \frac{1}{32} \cdot \left( c_0 \gamma \cdot \frac{\chisq{b}{b'}}{\exp(\ren{b}{b'})} \right)^2 \right) \cdot \sqrt{\exp(\ren{b}{b'})} \nonumber\\
      \leq&  \left( 1 - c_0^4 \gamma^4 / 32\right) \cdot \sqrt{\exp(\ren{b}{b'})}\label{eq:cgamma-4},
    \end{align}
    where (\ref{eq:cgamma-4}) uses the fact that $\frac{\chisq{b}{b'}}{\exp(\ren{b}{b'})} = \frac{\chisq{b}{b'}}{\chisq{b}{b'}+1} \geq c_0\gamma$ since we have assumed $\chisq{b}{b'} \geq c_0\gamma$. From (\ref{eq:exp-ay-4}), we have that
    \begin{align}
\E_{y \sim \BO_h^\t b} \left[ \exp \left( \frac{A(y)}{4} \right) \right] \leq & \sqrt{\E_{y \sim \BO_h^\t b} [\sqrt{F(y)}]} \leq (1 - c_0^4 \gamma^4/64) \cdot \exp \left( \frac{\ren{b}{b'}}{4} \right)\nonumber.
    \end{align}
    It follows that
    \begin{align}
\E_{y \sim \BO_h^\t b} \left[ \exp \left( \frac{A(y)}{4} \right) - 1 \right] \leq (1 - c_0^4 \gamma^4 /64) \cdot \left( \exp \left( \frac{\ren{b}{b'}}{4} \right) -1\right)\nonumber,
    \end{align}
    and thus, by Jensen's inequality, that
    \begin{align}
\E_{y \sim \BO_h^\t b} \left[ \sqrt{\exp(A(y)/4) - 1} \right] \leq (1 - c_0^4 \gamma^4/128) \cdot \sqrt{\exp(\ren{b}{b'}/4) - 1}\nonumber,
    \end{align}
as desired (note that $c_0^4/128 = 1/2^{39} > 1/2^{40}$).
    
  \paragraph{Case 2b.} If Case \xref{case:2a} does not hold, it holds that
    \begin{align}
\E_{y \sim \BO_h^\t b} \left[ |F(y) - \exp(\ren{b}{b'})| \right] \leq 2c_0 \gamma \cdot \chisq{b}{b'}\label{eq:fprime-expren}.
    \end{align}
    Define
    \begin{align}
      F'(y) := &\; \BO_h(y|b) \cdot (F(y) - \exp(\ren{b}{b'})) \nonumber\\
      =&\; \BO_h(y|b) \cdot \left( \sum_{x \in \MS} \frac{b(x) \BO_h(y|x)}{\BO_h(y|b)} \cdot \frac{b(x)}{b'(x)} - \sum_{x \in \MS} \frac{b(x)^2}{b'(x)} \right)\nonumber\\
      =&\; \sum_{x \in \MS} \BO_h(y|x) \cdot \frac{b(x)^2}{b'(x)} - \sum_{x \in \MS} \BO_h(y|b) \frac{b(x)^2}{b'(x)}\nonumber.
    \end{align}
    Thus, the vector $F' \in \BR^\MO$ may be written as
    \begin{align}
F' = \BO_h^\t \sum_{x \in \MS} (\delta_x - b) \cdot \frac{b(x)^2}{b'(x)}\nonumber,
    \end{align}
    where $\delta_x \in \BR^\MS$ is the indicator vector of $x$. Thus, by Assumption \ref{assumption:observability-intro},
    \begin{align}
\| F' \|_1 \geq \gamma \cdot \left\| \sum_{x \in \MS} (\delta_x - b) \cdot \frac{b(x)^2}{b'(x)} \right\|_1 = \gamma \cdot \sum_{z \in \MS} \left| \frac{b(z)^2}{b'(z)} - b(z) \cdot \exp(\ren{b}{b'}) \right|\label{eq:fprime-l1}.
    \end{align}
Note that $\| F'\|_1 = \E_{y \sim \BO_h^\t b} | F(y) - \exp(\ren{b}{b'})|$.  Thus, combining (\ref{eq:fprime-l1}) with (\ref{eq:fprime-expren}) gives that
    \begin{align}
2c_0\cdot \chisq{b}{b'} \geq \E_{x \sim b} \left| \frac{b(x)}{b'(x)} - \exp(\ren{b}{b'})\right|= \E_{x \sim b} \left| \left(\frac{b(x)}{b'(x)}-1 \right) - \chisq{b}{b'}\right|\nonumber.
    \end{align}
    By Lemma \ref{lem:g-chisq} and since $c_0 \leq 1/20$, it holds that $\tvd{b}{b'} \geq \min \{ \frac 18 \cdot \chisq{b}{b'}, \frac 17 \} \geq c_0\gamma/8$. Then
    \begin{align}
      \sqrt{\E_{y \sim \BO_h^\t b} \left[ \left( \frac{\BO_h(y|b')}{\BO_h(y|b)} \right)^{1/2} \right]} =& \sqrt{1 - \hell{\BO_h^\t b }{\BO_h^\t b'} }\nonumber\\
      \leq & \sqrt{1 - \frac 12 \cdot \tvd{\BO_h^\t b}{\BO_h^\t b'}^2} \label{eq:use-sq-hell}\\
      \leq & \sqrt{1 - \frac{\gamma^2}{2}\cdot \tvd{b}{b'}^2}\label{eq:hell-obs}\\
      \leq & 1 - c_0^2 \gamma^4 / 256\nonumber,
    \end{align}
    where (\ref{eq:use-sq-hell}) follows since $\hell{P}{Q} \geq \frac 12 \cdot \tvd{P}{Q}^2$ for all probability distributions $P,Q$, and (\ref{eq:hell-obs}) uses Assumption \ref{assumption:observability-intro}. From (\ref{eq:exp-ay-4}), we then have that
    \begin{align}
\E_{y \sim \BO_h^\t b} \left[ \exp \left( \frac{A(y)}{4} \right) \right] \leq & (1 - c_0^2 \gamma^4 / 256) \cdot \left( \E_{y \sim \BO_h^\t b}[F(y)]\right)^{1/4} = (1 - c_0^2 \gamma^4 / 256) \cdot \exp(\ren{b}{b'}/4)\nonumber.
    \end{align}
    As in Case \xref{case:2a}, it then follows that
    \begin{align}
\E_{y \sim \BO_h^\t b} \left[ \sqrt{ \exp(A(y)/4) -1} \right] \leq & (1 - c_0^2 \gamma^4/512) \cdot \sqrt{\exp(\ren{b}{b'}/4)-1}\nonumber,
    \end{align}
as desired (note that $c_0^2/512 = 1/2^{25}$).
\end{proof}

We immediately get the following corollary, from which we can prove Theorem~\ref{theorem:stability-renyi}.

\begin{corollary}\label{lemma:renyi-belief-update-contraction}
Let $b, b' \in \Delta^\MS$ with $b \ll b'$, and fix any $h \in [H-1]$. Then for any action $a$, with expectation over $y \sim (\bO_{h+1})^\t \bT_h(a) \cdot b$, $$\EE \sqrt{\exp(\ren{U_h(b;a,y)}{U_h(b';a,y)}/4) - 1} \leq \left(1 - \gamma^4/2^{40}\right)\sqrt{\exp(\ren{b}{b'}/4) - 1}.$$
\end{corollary}

\begin{proof}
Apply Lemma~\ref{lemma:renyi-bayes-update-contraction} and the data processing inequality (since $D_2$ is an $f$-divergence).
\end{proof}

\begin{proof}[\textbf{Proof of Theorem~\ref{theorem:stability-renyi}}]
  The proof closely follows that of Theorem \ref{theorem:stability-kl}. 
Fix some history $(a_{1:h-1},o_{2:h})$. We condition on this history throughout the proof. For $0 \leq n \leq t$ define the random variables $$b_{h+n} = \belief_{h+n}(a_{1:h+n-1},o_{2:h+n}),$$
$$b'_{h+n} = \sbelief_{h+n}(a_{h:h+n-1},o_{h+1:h+n}),$$
$$Y_n = \sqrt{\exp(D_2(b_{h+n}||b'_{h+n})/4) - 1}.$$

Then $$D_2(b_h||b'_h) = \log \EE_{x \sim b_h} \frac{b_h(x)}{b'_h(x)} \leq \log(S)$$ since $b'_h = \sbelief_h(\emptyset) = \Unif(\MS)$ (unless $h=1$, but then $b'_h = b_h$), so $Y_0 \leq \sqrt{\exp(D_2(b_h||b'_h))} \leq S$. Moreover, for any $0 \leq n < t$, we have
\begin{align*}
&\EE_{a_{h:h+n},o_{h+1:h+n+1} \sim \pi} Y_{n+1}
= \EE_{a_{h:h+n-1}, o_{h+1:h+n} \sim \pi}
\EE_{o_{h+n+1} \sim (\bO_{h+n+1})^\t \bT_{h+n}(\pi_{h+n-1}(a_{1:h+n-1},o_{2:h+n}))b_{h+n}} \\
& \sqrt{\exp(\ren{U_{h+n}(b_{h+n}; \pi_{h+n-1}(a_{1:h+n-1},o_{2:h+n}),o_{h+n+1})}{U_{h+n}(b'_{h+n}; \pi_{h+n-1}(a_{1:h+n-1},o_{2:h+n}),o_{h+n+1})}/4) - 1} \\
  &\leq \EE_{a_{h:h+n-1},o_{h+1:h+n} \sim \pi} (1 - \gamma^4/2^{40}) \sqrt{\exp(\ren{b_{h+n}}{b'_{h+n}}/4) - 1}\\
  &= (1 - \gamma^4/2^{40}) \cdot \EE_{a_{h:h+n-1}, o_{h+1:h+n} \sim \pi} Y_n
\end{align*}
by Corollary~\ref{lemma:renyi-belief-update-contraction}. By induction, we have that $$\EE_{a_{h:h+t-1},o_{h+1:h+t} \sim \pi} \sqrt{\exp(\ren{b_{h+t}}{b'_{h+t}}/4) - 1} \leq (1-\gamma^4/2^{40})^t S \leq \epsilon$$ by choice of $t$. It follows from Lemma~\ref{lemma:d2-to-tv} that $$\EE_{a_{h:h+t-1},o_{h+1:h+t} \sim \pi} \norm{b_{h+t} - b'_{h+t}}_1 \leq 4\epsilon.$$
Taking expectation over the history $(a_{1:h-1},o_{2:h})$ completes the proof.
\end{proof}

\section{Planning with Short Memory} \label{section:planning}

\begin{algorithm}
    \caption{\textsc{Short-Memory Planning}}\label{alg:smp}
    \label{alg:apvi}
    \begin{algorithmic}[1]
        \Procedure{SMP}{$(H,\MS,\MA,\MO,b_1, R,\bT,\bO)$, $L$}
            \State $\hat{V}_H(a_{\max(1,H-L):H-1}, o_{\max(2,H-L+1):H}) \gets 0$ for all $a,o$
            \State $h \gets H-1$
            \While{$h \neq 0$}
                \State $b' \gets b(a_{\max(1,h-L):h-1},o_{\max(2,h-L+1):h})$
                \State $\hat{Q}_h((a_{\max(1,h-L):h-1},a),o_{\max(2,h-L+1):h}) \gets$
                
                \qquad \qquad$\EE_{y \sim (\bO_{h+1})^\t \bT_h(a) b'}\left[R_{h+1}(y) + \hat{V}_{h+1}((a_{\max(1,h+1-L):h-1},a),(o_{\max(2,h-L+2):h},y))\right]$
                \State $\hat{\pi}_h(a_{\max(1,h-L):h-1},o_{max(2,h-L+1):h}) \gets \argmax_{a \in \MA} \hat{Q}_h((a_{\max(1,h-L):h-1},a),o_{\max(2,h-L+1):h})$
                \State $\hat{V}_h(a_{\max(1,h-L):h-1},o_{\max(2,h-L+1):h}) \gets \max_{a \in \MA} \hat{Q}_h((a_{\max(1,h-L):h-1},a),o_{\max(2,h-L+1):h})$
                \State $h \gets h - 1$
            \EndWhile
            \State \textbf{return} $\hat{\pi}$
        \EndProcedure
    \end{algorithmic}
\end{algorithm}

In this section, we formally describe the short-memory planning algorithm (Algorithm~\ref{alg:smp}), and show that it achieves $\epsilon$-suboptimality on observable POMDPs in quasipolynomial time, thereby proving Theorem~\ref{thm:main-intro}. Given the description of a POMDP and a history length $L \in \BN$, Algorithm~\ref{alg:smp} inductively computes a (quasi-succinct description of) policy $\hat{\pi}$ as follows.

\begin{definition}
Define $\hat{V}_H = 0$. For $1 \leq h < H$, let 
\begin{align*}&\hat{Q}_h((a_{\max(1,h-L):h-1},a),o_{\max(2,h-L+1):h})  \\
&\qquad= \EE_{y \sim (\bO_{h+1})^\t \bT_h(a) \sbelief_h(a_{\max(1,h-L):h-1},o_{\max(2,h-L+1):h})} \left[R_{h+1}(y) + \hat{V}_{h+1}((a_{1:h-1},a),(o_{2:h},y))\right].\end{align*}
$$\hat{V}_h(a_{\max(1, h-L):h-1}, o_{\max(2, h-L+1):h}) = \max_{a \in \MA} \hat{Q}_h((a_{\max(1,h-L):h-1},a),o_{\max(2,h-L+1):h}).$$
$$\hat{\pi}_h(a_{1:h-1},o_{2:h}) = \argmax_{a \in \MA} \hat{Q}_h((a_{\max(1,h-L):h-1},a),o_{\max(2,h-L+1):h}).$$
\end{definition}

Recall that the value function of the policy $\hat \pi$, denoted $V_h^{\hat \pi}(a_{1:h-1}, o_{2:h})$ (for $h \in [H]$) is defined in Definition \ref{def:pol-valfn}. Moreover, the value function of the optimal policy is described in Proposition~\ref{prop:optimal-valfn}. We are interested in bounding the suboptimality of $\hat{\pi}$, which is $V^*_1(\emptyset) - V^{\hat{\pi}}_1(\emptyset)$.

In Lemmas \ref{lemma:vstar-shortv} and \ref{lemma:shortv-vpi} below, we relate $V_h^{\hat{\pi}}$ to $V_h^*$ via the approximate value function $\hat{V}_h$.

\begin{lemma}
  \label{lemma:vstar-shortv}
Suppose that $$\EE_{a_{1:h-1},o_{2:h} \sim \pi} \norm{\belief_h(a_{1:h-1},o_{2:h}) - \sbelief_h(a_{\max(1,h-L):h-1},o_{\max(2,h-L+1):h})}_1 \leq \epsilon$$ for every policy $\pi$ and $h \in [H]$. Then for any policy $\pi$, and any $1 \leq h \leq H$, it holds that $$\EE_{a_{1:h-1},o_{2:h} \sim \pi} V^*_h(a_{1:h-1},o_{2:h}) - \hat{V}_h(a_{\max(1,h-L):h-1},o_{\max(2,h-L+1):h}) \leq \epsilon H(H-h).$$
\end{lemma}

\begin{proof}
We proceed by induction on $h$. If $h = H$ then both functions are identically zero. Let $h<H$ and suppose that the claim holds for $h+1$. Fix a policy $\pi$. Then for any action/observation sequence $a_1,\dots,o_h$, if $a^*$ is the optimal next action for the optimal policy, we have
\begin{align*}
&V_h^*(a_{1:h-1},o_{2:h}) - \hat{V}_h(a_{\max(1,h-L):h-1},o_{\max(2,h-L+1):h}) \\
&\leq Q_h^*((a_{1:h-1},a^*),o_{2:h}) - \hat{Q}_h((a_{\max(1,h-L):h-1},a^*),o_{\max(2,h-L+1):h}) \\
&\leq (H-h)\norm{\belief_h(a_{1:h-1},o_{2:h}) - \sbelief_h(a_{\max(1,h-L):h-1},o_{\max(2,h-L+1):h})}_1 \\
&\quad+ \EE_{y\sim (\bO_{h+1})^\t\bT_h(a^*)b(a_{1:h-1},o_{2:h})}\left[V^*_{h+1}((a_{1:h-1},a^*),(o_{2:h},y)) - \hat{V}_{h+1}((a_{\max(1,h-L+1):h-1},a^*),(o_{\max(2,h-L+2):h},y))\right].
\end{align*}
Define a policy $\pi'$ which follows $\pi$ until time $h$, and then follows the optimal policy. Then for $a_{1:h-1},o_{2:h} \sim \pi$ and $a^* = \pi^*(a_{1:h-1},o_{2:h})$ and $y \sim (\bO_{h+1})^\t \bT_h(a^*)b(a_{1:h-1},o_{2:h})$, we have that $(a_{1:h-1},a^*), (o_{2:h},y) \sim \pi'$. Thus, by the induction hypothesis and the assumption,
\begin{align*}
&\EE_{a_{1:h-1},o_{2:h} \sim \pi} V_h^*(a_{1:h-1},o_{2:h}) - \hat{V}_h(a_{\max(1,h-L):h-1},o_{2:h}) \\
&\leq (H-h)\EE_{a_{1:h-1},o_{2:h} \sim \pi} \norm{\belief_h(a_{1:h-1},o_{2:h}) - \sbelief_h(a_{\max(1,h-L):h-1},o_{\max(2,h-L+1):h)}}_1 \\
&\qquad + \EE_{(a_{1:h-1},a^*),(o_{2:h},y) \sim \pi'} \left[V^*_{h+1}((a_{1:h-1},a^*),(o_{2:h},y)) - \hat{V}_{h+1}((a_{\max(1,h-L+1):h-1},a^*),(o_{\max(2,h-L+2):h},y))\right] \\
&\leq (H-h)\epsilon + H(H-h-1)\epsilon \\
&\leq H(H-h)\epsilon
\end{align*}
as desired.
\end{proof}

\begin{lemma}\label{lemma:shortv-vpi}
Suppose that $$\EE_{a_{1:h-1},o_{2:h} \sim \pi} \norm{\belief_h(a_{1:h-1},o_{2:h}) - \sbelief_h(a_{\max(1,h-L):h-1},o_{\max(2,h-L+1):h})}_1 \leq \epsilon$$ for every policy $\pi$ and $h \in [H]$. Then for any policy $\pi$, and any $1 \leq h \leq H$, it holds that $$\EE_{a_{1:h-1},o_{2:h} \sim \pi} \hat{V}_h(a_{\max(1,h-L):h-1},o_{\max(2,h-L+1):h}) - V^{\hat{\pi}}_h(a_{1:h-1},o_{2:h}) \leq \epsilon H(H-h).$$ 
\end{lemma}

\begin{proof}
We proceed by induction on $h$. If $h = H$ then both functions are identically zero. Let $h < H$ and suppose that the claim holds for $h+1$. Fix a policy $\pi$. Then for any action/observation sequence $a_{1:h-1},o_{2:h}$, if $\hat{a} = \hat{\pi}(a_{1:h-1},o_{2:h})$, then we have
\begin{align*}
& \hat{V}_h(a_{\max(1,h-L):h-1},o_{\max(2,h-L+1):h}) - V^{\hat{\pi}}_h(a_{1:h-1},o_{2:h}) \\
&= \EE_{y \sim (\bO_{h+1})^\t \bT_h(\hat{a}) \sbelief_h(a_{\max(1,h-L):h-1},o_{\max(2,h-L+1):h})} \left[R_{h+1}(y) + \hat{V}_{h+1}((a_{\max(1,h-L+1):h-1},\hat{a}),(o_{\max(2,h-L+2):h},y))\right] \\
&\quad - \EE_{y \sim (\bO_{h+1})^\t \bT_h(\hat{a})b(a_{1:h-1},o_{2:h})}\left[R_{h+1}(y) + V^{\hat{\pi}}_{h+1}((a_{1:h-1},\hat{a}),(o_{2:h},y))\right] \\
&\leq (H-h)\norm{\sbelief_h(a_{\max(1,h-L):h-1},o_{\max(2,h-L+1):h}) - \belief(a_{1:h-1},o_{2:h})}_1 \\
&\quad+ \EE_{y \sim (\bO_{h+1})^\t \bT_h(\hat{a})\belief(a_{1:h-1},o_{2:h})}\left[\hat{V}_{h+1}((a_{\max(1,h-L+1):h-1},\hat{a}),(o_{\max(2,h-L+2):h},y)) - V^{\hat{\pi}}((a_{1:h-1},\hat{a}),(o_{2:h},y))\right].
\end{align*}
Define a policy $\pi'$ which follows $\pi$ until time $h$, and then follows policy $\hat{\pi}$. Then for $a_{1:h-1},o_{2:h} \sim \pi$ and $\hat{a} = \hat{\pi}(a_{1:h-1},o_{2:h})$ and $y \sim (\bO_{h+1})^\t \bT_h(\hat{a})b(a_{1:h-1},o_{2:h})$, we have that $(a_{1:h-1},\hat{a}),(o_{2:h},y) \sim \pi'$. Thus, by the induction hypothesis and the assumption,
\begin{align*}
&\EE_{a_{1:h-1},o_{2:h} \sim \pi} \hat{V}_h(a_{\max(1,h-L):h-1},o_{\max(2,h-L+1):h}) - V^{\hat{\pi}}_h(a_{1:h-1},o_{2:h}) \\
&\leq (H-h)\EE_{a_{1:h-1},o_{2:h} \sim \pi}\norm{\sbelief_h(a_{\max(1,h-L):h-1},o_{\max(2,h-L+1):h}) - \belief(a_{1:h-1},o_{2:h})}_1 \\
&\quad + \EE_{(a_{1:h-1},\hat{a}),(o_{2:h},y) \sim \pi'} \left[\hat{V}_{h+1}((a_{\max(1,h-L+1):h-1},\hat{a}),(o_{\max(2,h-L+2):h},y)) - V^{\hat{\pi}}((a_{1:h-1},\hat{a}),(o_{2:h},y))\right] \\
&\leq (H-h)\epsilon + H(H-h-1)\epsilon \\
&\leq H(H-h)\epsilon
\end{align*}
as desired.
\end{proof}

We can now prove our main result.

\begin{proof}[\textbf{Proof of Theorem~\ref{thm:main-intro}}]
Given any POMDP on $S$ states and any $\epsilon,\gamma > 0$, let $\hat{\pi}$ be the policy produced by Algorithm~\ref{alg:smp}, with window length $L \geq C\gamma^{-4}\log(SH/\epsilon)$ chosen so that Theorem~\ref{theorem:stability-renyi} applies with bound $\epsilon/(2H^2)$. By Theorem~\ref{theorem:stability-renyi}, we have that $$\EE_{a_{1:h-1},o_{2:h} \sim \pi} \norm{\belief_h(a_{1:h-1},o_{2:h}) - \sbelief_h(a_{\max(1,h-L):h-1},o_{\max(2,h-L+1):h})}_1 \leq \frac{\ep}{2H^2}$$ for any policy $\pi$ and $h > L$. If $h \leq L$, then $\sbelief_h(a_{\max(1,h-L):h-1},o_{\max(2,h-L+1):h}) = \belief_h(a_{1:h-1},o_{2:h})$ by definition, so the bound still holds. Thus, we can apply Lemmas~\ref{lemma:vstar-shortv} and~\ref{lemma:shortv-vpi} to get $$\EE_{a_{1:h-1},o_{2:h} \sim \pi} V^*_h(a_{1:h-1},o_{2:h}) - V^{\hat{\pi}}_h(a_{1:h-1},o_{2:h}) \leq \epsilon$$ for any $h \in [H]$. In particular, taking $h = 1$, we get that $V^* - V^{\hat{\pi}} \leq \epsilon$ as desired. Moreover, the time complexity of the algorithm is $H(OA)^L \cdot \poly(S,A,O)$; by choice of $L$, the quasipolynomial time complexity bound follows.
\end{proof}

By using Theorem \ref{theorem:stability-kl} in place of Theorem \ref{theorem:stability-renyi}, we can also prove the following result, which achieves better dependence on $\gamma$ but worse dependence on $H,S, 1/\ep$ than Theorem \ref{thm:main-intro}:
\begin{theorem}
  \label{thm:main-kl}
Let $\epsilon,\gamma > 0$. There is an algorithm which given the description of a $\gamma$-observable POMDP with $S$ states, $A$ actions, $O$ observations, and horizon length $H$, outputs an $\epsilon$-suboptimal policy and has time complexity $H(OA)^{C\log(SH/\epsilon)\log(H \log(S)/\ep)/\gamma^2}$ for some universal constant $C>0$.
\end{theorem}
\begin{proof}
The proof is identical to that of Theorem \ref{thm:main-intro}, except Theorem \ref{theorem:stability-renyi} is used in place of Theorem \ref{theorem:stability-kl}, with window size $L = C \gamma^{-2} \log(HS/\ep) \log(H\log(S)/\ep)$, where the constant $C$ is chosen so that Theorem \ref{theorem:stability-kl} applies with bound $\ep/(2H^2)$. 
\end{proof}

\section{Lower bounds}\label{section:lower-bounds}

\subsection{Quasipolynomial lower bound for observable POMDPs}

In this section we prove that quasipolynomial time is necessary to solve the planning problem in observable POMDPs, under the Exponential Time Hypothesis:

\begin{conjecture}[ETH \cite{impagliazzo2001complexity}]
  \label{con:eth}
  There is no $2^{o(n)}$-time algorithm which can determine whether a given 3SAT formula on $n$ variables is satisfiable.
\end{conjecture}

Let $\phi$ be a 3SAT formula on $n$ variables and $m$ clauses. Fix some $\gamma > 0$ satisfying $1/n \leq \gamma \leq 1/2$. We next construct a $\gamma$-observable POMDP $M_\phi$ based on $\phi$ as follows. At a high level, a plan for the POMDP $M_{\phi,\gamma}$ specifies, for each of some number $T$ of independent \emph{trials}, an assignment of the $n$ variables of $\phi$. For each trial, $M_{\phi,\gamma}$ chooses one of the $m$ clauses uniformly at random (as enconded in $M_{\phi,\gamma}$'s transition matrices), and the subsequent  transitions of that trial check whether the chosen assignment satisfies that clause. The value of any policy for $M_{\phi,\gamma}$ lies in $[0,1]$; in order for the value of some policy to be close to 1, it is necessary that the assignments chosen by that policy satisfy all of the $T$ trials with high probability. As we will show, by making $T$ large enough compared to $\gamma$, we can ensure that if (and only if) there is a policy that satisfies all $T$ trials with high probability, then $\phi$ is satisfiable. The intuition behind this fact is that for a small fraction (dependent on $\gamma$) of the trials, the observations will not be informative about the clause for the respective trial, meaning that, if $\phi$ is not satisfiable, any truth assignment fails the trial with probability at least $1/m$ (Lemma \ref{lemma:fail-lb}), which may be amplified by repetition. Conversely, if $\phi$ is satisfiable, its satisfying assignment may easily be seen to have value 1 (Lemma \ref{lem:succ-one}). The formal description of $M_{\phi,\gamma} = (\MS, \MA, \MO, H, \BP, \BO, R)$ given $\phi$ is as follows:

\nc{\sat}{{\text{satisfied}}}
\nc{\fail}{{\text{failed}}}
\begin{enumerate}
\item Set $T := 2n^3 \cdot \exp(2\sqrt{\gamma n})$.
    \item The state space is $$\mathcal{S} = [T+1] \times \{0,1\} \times [m] \times [\sqrt{n/\gamma}] \times \{0,1\}$$
    where the $[T]$ denotes $T$ independent successive trials, the first $\{0,1\}$ denotes whether any trials so far have failed, $[m]$ denotes the random clause picked for the current trial, the $[\sqrt{n/\gamma}]$ denotes the step of the current trial, and the second $\{0,1\}$ denotes whether the clause in this trial has been satisfied yet by the variable assignments given by the policy. 
    \item The action space is $$\mathcal{A} = \{0,1\}^{\sqrt{\gamma n}},$$ interpreted as an assignment to a block of $\sqrt{\gamma n}$ variables.
    \item The horizon is $H = T \cdot \sqrt{n/\gamma}$.
    \item The observation of a state $(t, b_{\text{failed}}, i, j, b_{\text{satisfied}})$ is:
\begin{itemize}
    \item With probability $\gamma$, the whole state $(t, b_{\text{failed}}, i, j, b_{\text{satisfied}})$
    \item With probability $1-\gamma$, a uniformly random state from $\MS$.
    \end{itemize}
    In particular, the space of observations satisfies $\MO = \MS$ and all emission matrices $\BO_h$ are the matrices from item \ref{it:gamma-identity} of Example \ref{ex:simple-observable}.
    
    \item The initial state distribution $b_1$ is uniform on the states $$\{1\} \times \{0\} \times [m] \times \{1\} \times \{0\}.$$
    
    \item For each state $(t, b_{\text{failed}}, i, j, b_{\text{satisfied}}) \in \mathcal{S}$ with $t \leq T$, and each action $a \in \{0,1\}^{\sqrt{\gamma n}}$, we define the transition to the next state as follows. For notational convenience, let $g \in \{0,1\}$ be the indicator for the event that the assignment from $a$ to variables $\{(j-1)\sqrt{n\gamma} + 1, \dots, j\sqrt{n\gamma}\}$ satisfies clause $i$ (i.e. some positive variable in clause $i$ is assigned $1$ by $a$, or some negative variable in clause $i$ is assigned $0$ by $a$). Then:

    \begin{itemize}
        \item If $j < \sqrt{n/\gamma}$, transition to state $$(t, b_\text{failed}, i, j+1, b_\text{satisfied} \lor g).$$
        \item Otherwise, draw $i' \sim \Unif([m])$ and transition to state $$(t+1, b_\text{failed} \lor \lnot (b_\text{satisfied} \lor g), i', 1, 0).$$
    \end{itemize}
    
    The states $(t, b_{\text{failed}}, i, j, b_{\text{satisfied}})$ with $t = T+1$ are terminal states. 
    
    \item The reward of every non-terminal state is $0$. The reward of a terminal state $(T+1, b_\text{failed}, i, 1, b_\text{satisfied})$ is precisely $1 - b_\text{failed}$.
\end{enumerate}

    The respective spaces have the following sizes: $A = 2^{\sqrt{n\gamma}}$, $S = 4m\sqrt{n/\gamma} \cdot (T+1) \leq 8m \sqrt{n/\gamma} \cdot T$, $O=S$, $H = (T+1)\sqrt{n/\gamma} \leq 2 T \sqrt{n/\gamma}$.

\begin{lemma}
  \label{lem:succ-one}
Suppose that $\phi$ is satisfiable. Then there is a policy achieving value 1.
\end{lemma}
\begin{proof}
Let $x \in \{0,1\}^n$ denote a satisfying assignment. Let $\pi$ be the policy that, at timestep corresponding to index $j \in [\sqrt{n/\gamma}]$, sets $a_k = x_{(j-1)\sqrt {n\gamma} + k}$ for $1 \leq k \leq \sqrt {n\gamma}$, and plays the action $({a_1}, \ldots, a_{\sqrt{n\gamma}})$. Since any clause $i$ is satisfied by some variable, for each trial there is some step in which $g = 1$, so at the end of each trial we always have $b_\text{satisfied} \lor g = 1$, so we never change $b_\text{failed}$ from its initial value of $0$, which means that the total reward incurred is $1$.
\end{proof}

\begin{lemma}\label{lemma:fail-lb}
Suppose that $\phi$ is unsatisfiable. Then for any policy $\pi$, for any trial $1 \leq t \leq T$, the probability that the state reaches $b_\text{failed} = 1$ by the end of the trial is at least $$\frac{1}{m} \cdot (1-\gamma)^{\sqrt{n/\gamma}},$$ conditioned on all previous trials.
\end{lemma}

\begin{proof}
Fix any policy $\pi$ and any trial $t$. Condition on the randomness of the first $t-1$ trials. At each step $j$ of trial $t$, let $f_j$ be the $\text{Ber}(1-\gamma)$ random variable where $f_j = 1$ if the observation is uniformly random and $f_j = 0$ otherwise. We can consider $f_1,\dots,f_{\sqrt{n/\gamma}}$ as all being drawn at the beginning of the trial. Then conditioned on the event $E = \{f_i = 1 \quad \forall i\}$, the clause is still uniformly random, and the observations are all uniformly random, and therefore independent of the clause. Thus, the actions are independent of the clause, so the clause is uniformly distributed even after conditioning on the actions. But then since $\phi$ is unsatisfiable, for any action sequence the probability of failure is at least $1/m$. This holds conditioned on event $E$, which has probability $(1-\gamma)^{\sqrt{n/\gamma}}$. Thus, the unconditional probability of failure is at least $\frac{1}{m} (1-\gamma)^{\sqrt{n/\gamma}}$ as claimed.
\end{proof}

We say that a function $\bar \gamma : \BN \ra \BR_{ > 0}$ is \emph{bounded decay} if $\bar \gamma$ is non-increasing, $\bar \gamma(K) \leq 2 \cdot \bar \gamma(K+1)$ for all $K$, and $\bar \gamma(K) \in [K^{-1/10}, 1/2]$ for all $K$.
\begin{theorem}
  \label{thm:eth-lb}
Suppose there is an algorithm and a function $\bar \gamma : \BN \ra \BR_{ > 0}$ of bounded decay, satisfying the following: for any $K \in \BN$, for $\gamma := \bar \gamma(K)$, given any $\gamma$-observable POMDP with $\max\{S, A, O, H\} \leq K$, the algorithm determines the value of the optimal policy up to error $1/10$ and runs in time $K^{o \left( \frac{\log K}{\gamma} \right)}$. Then Conjecture \ref{con:eth} is false.
\end{theorem}
\begin{proof}
  Fix any 3-SAT instance $\phi$ with $n$ variables and $m$ clauses. Without loss of generality, we have $m \leq n^3$. Consider some $1/n \leq \gamma \leq 1/2$, to be specified exactly below. Construct the POMDP $M_{\phi,\gamma}$ as described above; recall that we set $T = 2n^3 \cdot \exp(2\sqrt{\gamma n})$.
  By Lemma \ref{lem:succ-one}, if $\phi$ is satisfiable, then the value of the POMDP $M_{\phi,\gamma}$ is 1.

  On the other hand, suppose $\phi$ is not satisfiable. The value of the POMDP is the probability it reaches a terminal state $(t+1,b_\fail, i,1,b_\sat)$ with $b_\fail = 0$, which, by Lemma \ref{lemma:fail-lb}, is bounded above by
  $$
  (1 - (m^{-1} \cdot (1-\gamma)^{\sqrt{n/\gamma}}))^T \leq \left( 1 - m^{-1} \cdot \exp(-2 \sqrt{\gamma n}) \right)^T,
  $$
  where the inequality follows since $1-\gamma \geq \exp(-2\gamma)$ for $\gamma \leq 1/2$. Note that for all reals $x < 1$, we have $1/x > \frac{1}{\ln(1/(1-x))}$. Thus, as long as
  \begin{align}
T \geq \frac{2}{m^{-1} \cdot \exp(-2\sqrt{\gamma n})} \geq \frac{\ln(2)}{\ln \left( \frac{1}{1-m^{-1} \cdot \exp(-2 \sqrt{\gamma n})} \right)}\nonumber,
  \end{align}
  we have that the optimal value of $M$ is at most $1/2$. Note that the size of the respective parameters of $M$ are bounded as follows (using that $1/n \leq \gamma$ and $m \leq n^3$):
  \begin{align}
A = 2^{\sqrt{\gamma n}}, \qquad H = (T+1) \sqrt{n/\gamma} \leq O(n^4 \cdot \exp(2\sqrt{\gamma n})), \qquad \max\{S, O \} \leq O(m \sqrt{n/\gamma} T) \leq O(n^7 \cdot \exp(2 \sqrt{\gamma n}))\nonumber.
  \end{align}
  In particular, for some sufficiently large constant $C > 0$, if we take $K = \lfloor C \cdot n^7 \cdot \exp(2 \sqrt{\gamma n})\rfloor$, we have that $\max\{S,A,H,O\} \leq K$. Recall the non-increasing function $\bar\gamma(\cdot)$ from the theorem statement; now we specify that $$\gamma = \inf \{x: \bar{\gamma}(\lfloor Cn^7 \cdot \exp(2\sqrt{x n}) \rfloor)\} \leq x\}.$$ Since $\bar{\gamma}$ is of bounded decay, the set contains $x=1/2$, so $\gamma$ is well-defined and at most $1/2$. Moreover, if $x < 1/n$ then $$\bar{\gamma}(\lfloor Cn^7 \cdot \exp(2\sqrt{xn})\rfloor) \geq \bar{\gamma}(\lfloor Cn^7 \cdot \exp(2)\rfloor) \geq \Omega(n^{-7/10}) > 1/n = x$$ for sufficiently large $n$, where the first inequality uses that $\bar{\gamma}$ is non-increasing. Thus, $\gamma \in [1/n, 1/2]$ as required.
  
    Finally, by minimality of $\gamma$, either $\gamma = \bar{\gamma}(\lfloor Cn^7 \cdot \exp(2\sqrt{xn})\rfloor)$ or $Cn^7 \cdot \exp(2\sqrt{\gamma n})$ is some integer $k$. In the latter case, for any sufficiently small $\epsilon > 0$, we have $$\bar{\gamma}(k) \geq \frac{1}{2}\bar{\gamma}(k-1) = \frac{1}{2}\bar{\gamma}(\lfloor Cn^7 \cdot \exp(2\sqrt{(\gamma-\epsilon)n})\rfloor) > \frac{1}{2}(\gamma - \epsilon)$$ and $$\gamma + \epsilon \geq \bar{\gamma}(\lfloor Cn^7 \cdot \exp(2\sqrt{(\gamma+\epsilon)n})\rfloor) = \bar{\gamma}(k).$$
    Thus, in either case, it holds that $\gamma/2 \leq \bar{\gamma}(K) \leq \gamma$ where $K = \lfloor Cn^7 \cdot \exp(2\sqrt{\gamma n})\rfloor$. In particular, the constructed POMDP has $\max(S,A,H,O) \leq K$ and is $\gamma$-observable (Example~\ref{ex:simple-observable}), and thus $\bar{\gamma}(K)$-observable. Hence, by the assumption in the theorem statement, the algorithm can determine the value of the optimal policy up to error $1/10$, and therefore determine whether $\phi$ is satisfiable, in time
  \begin{align}
K^{o(\log(K)/\bar{\gamma}(K))} \leq \exp\left( o \left(\frac{(2  \sqrt{\gamma n} + 7 \ln(n))^2}{\gamma} \right)\right) \leq \exp \left( o\left( \frac{8 \gamma n}{\gamma} + \frac{98 \ln(n)^2}{\gamma} \right) \right) \leq \exp(o(n)) \label{eq:eth-contradiction},
  \end{align}
  where the first inequality uses that $\bar{\gamma}(K) \geq \gamma/2$, and the final inequality follows as long as $\gamma \geq \ln^2(n) /n$. This holds because, for $K' = C \cdot n^7 \cdot \exp(2 \sqrt{\gamma n})$, the inequality $\gamma \geq \bar{\gamma}(K') \geq (K')^{-1/10}$ implies that
  \begin{align}
\ln(1/\gamma) \leq \ln(1/\bar{\gamma}(K')) \leq \frac{1}{10} \cdot \left( \ln(C) + 7 \ln(n) + 2 \sqrt{\gamma n}\right)\nonumber,
  \end{align}
  and if $\gamma < \ln^2(n)/n$, then it would follow that
  \begin{align}
\ln(n) - 2 \ln \ln(n) \leq \ln(1/\gamma) \leq \frac{1}{10} \cdot \left( \ln(C) + 7 \ln(n) + 2 \ln(n) \right) = \frac{\ln(C) + 9 \ln(n)}{10}\nonumber,
  \end{align}
  which is a contradiction for $n$ sufficiently large. We conclude that the algorithm contradicts Conjecture~\ref{con:eth}.
\end{proof}
We remark that the lower bound of Theorem \ref{thm:eth-lb} uses POMDPs $M_{\phi,\gamma}$ for which all observation matrices are of the type in item \ref{it:gamma-identity} of Example \ref{ex:simple-observable}. With minimal changes to the proof it may readily be seen the same lower bound holds instead using observation matrices of the type in item \ref{it:obs-null} of Example \ref{ex:simple-observable} (in particular, it only remains to note that in Lemma \ref{lemma:fail-lb} the actions are still independent of the clause under the event $E$).

\subsection{Lower bound for weak observability}


In this section we give a proof of Proposition~\ref{prop:weak-observability-lower-bound}. Concretely, we show that SAT instances can be embedded in weakly observable POMDPs; the construction is similar to that of the previous section (though simpler) so we omit some details. 

\begin{proof}[Proof of Proposition~\ref{prop:weak-observability-lower-bound}]
Given a SAT instance with $n$ variables $x_1, \ldots, x_n$ and $m$ clauses, we construct a POMDP with horizon length $H = n$ and where the state space is given by $\MS := [2m] \times [n] \times \{0,1\}$. For a state $(j, i, b) \in \MS$, the index $j \in [2m]$ denotes a clause; we pair up the elements of $[2m]$ as $(1,2), (3,4), \ldots, (2m-1,2m)$, so that each clause can be represented by two possible values of $j$. Furthermore, the bit $b$ keeps track of whether the current clause has been satisfied yet, and the index $i \in [n]$ keeps track of the step number (each step $i$ corresponds to a variable $x_i$). At step $n$ the POMDP has reward $1$ if the clause has been satisfied and $0$ otherwise; all other rewards are 0. The action space is $\MA := \{0,1\}$; at step $i$, an action $a_i \in \{0,1\}$ corresponds to an assignment for variable $x_i$, and the state transitions accordingly (in particular, the bit $b$ is set to 1 if the action leads to an assignment of $x_i$ that satisfies the chosen clause). The initial distribution over states is uniform over $[2m] \times [1] \times \{0\}$ (corresponding to a clause being chosen at random).

We proceed to describe the observation space and the emission matrix. Without loss of generality assume that $m$ is one less than power of $2$, and let $H \in \{-1,1\}^{m \times (m+1)}$ be the $(m+1) \times (m+1)$ Hadamard matrix with the first row (which has all 1's) removed. Thus each row of $H$ has $(m+1)/2$ 1's and $(m+1)/2$ $-1$'s. For $j \in [m]$ let $H_j$ denote the $j$th row of $H$. For states $(2j-1,i,b)$ and $(2j,i,b)$ (where $2j-1$ and $2j$ are the pair corresponding to clause $j$, and $i$ is the current step, and $b$ is the bit indicating whether clause $j$ has been satisfied), we define the observation distributions as follows. The observation for state $(2j-1,i,b)$ is $(X, i, b)$ where $X$ is uniform on the positive entries of $H_j$, and the observation for state $(2j,i,b)$ is $(Y,i,b)$ where $Y$ is uniform on the negative entries of $H_j$. Overall, we have $\MO = [m+1] \times [n] \times \{0,1\}$.

For any vectors $u,v \in \{-1,+1\}^m$ with $\sum u_i = \sum v_i = 0$, if $X$ and $Y$ are uniform on the positive entries of $u$ and $v$ respectively, then $$\TV(X,Y) = \frac{1}{2m} \sum_i |u_i - v_i| = \frac{1}{2m} \sum_i (1 - u_i v_i) = \frac{1}{2} - \frac{u\cdot v}{2m}.$$
But then for any for any $j,j'$ with $j \neq j'$, and any $c,c' \in \{0,1\}$, the total variation distance between observation distributions of states $(2j+c,i,b)$ and $(2j'+c',i,b)$ is $1/2$, since all rows of the Hadamard matrix are orthogonal, and negating a row does not change this fact. And for any $j$, the total variation distance between observation distributions of states $(2j,i,b)$ and $(2j+1,i,b)$ is $1$. We conclude that this family of observation distributions is $\gamma$-weakly-observable with $\gamma = 1/2$.

If the given SAT instance is satisfiable by some choice $x_1, \ldots, x_n \in \{0,1\}^n$, then choosing action $x_i$ at step $i$ gives a policy with value 1. On the other hand, if the SAT instance is not satisfiable, then no policy can have value greater than $1-1/m$: if such a policy $\pi$ existed, then consider the random assignment $x_1, \ldots, x_n$ defined as follows. Let $Z_1, \ldots, Z_n$ be drawn uniformly on $[m+1]$, and set $x_i$ to be the action taken by $\pi$ on step $i$ given that the previous observations at steps $i' < i$ are $(Z_{i'}, i', 0)$. For $j$ drawn uniformly from $[m]$ (independently of the $x_i$), it is straightforward to see that the probability (over both $j$ and $x_1, \ldots, x_n$) that $x_1, \ldots, x_n$ satisfies the $j$th clause is equal to the value of $\pi$, since the observations $(Z_{i'}, i', 0)$ are distributed identically to those in the actual POMDP under the policy $\pi$, up until the step $i$ at which the chosen clause of the POMDP is satisfied (at which point the reward is guaranteed to be 1). But we have assumed the value of $\pi$ to be greater than $1-1/m$. Thus $x_1, \ldots, x_n$ satisfy all clauses with positive probability, contradicting that the SAT instance is not satisfiable.

Thus the value of this POMDP encodes satisfiability of the original SAT instance (so that an $\epsilon$-approximation POMDP planning algorithm with $\epsilon < 1/m$ determines satisfiability), and the size of the POMDP is polynomial in $n$ and $m$; specifically, $SAOH \leq O(n^3 m^2)$. Thus, for some absolute constant $c>0$, a POMDP planning algorithm which $\epsilon$-additively approximates the value of $1/2$-weakly observable POMDPs in time $\exp(O((SAOH/\epsilon)^c))$ would contradict the Exponential Time Hypothesis.
\end{proof}

\subsection{Lower bound for belief contraction}

In this section we show that the quadratic dependence on $\gamma$ achieved in our simpler belief stability result (Theorem~\ref{theorem:stability-kl}) is optimal.

\begin{proposition}\label{prop:belief-contraction-lower-bound}
There is a constant $c > 0$ such that the following holds. Let $\gamma \in (0,1/10)$ and $H \in \BN$. There is a $\gamma$-observable POMDP with horizon length $H$, and a policy $\pi$, such that for all $1 < h < h+t \leq H$, it holds that $$\EE_{a_{1:h+t-1},o_{2:h+t} \sim \pi} \norm{\belief_{h+t}(a_{1:h+t-1},o_{2:h+t}) - \sbelief_{h+t}(a_{h:h+t-1},o_{h+1:h+t})}_1 \geq c \cdot \exp(-16\gamma^2 t).$$
\end{proposition}

\begin{proof}
We define a POMDP with state space and observation space $\mathcal{S} = \mathcal{O} = \{0,1\}$. Let the action space $\MA$ be arbitrary, but with trivial transitions, i.e. $\bT_h(a)$ is the identity operator for all $a \in \MA$ and $h \in [H]$. Define the initial state distribution to be $b_1 = (1,0)$. For $2 \leq h \leq H$, define $$\bO_h = \bO = \begin{bmatrix} 1/2 + \gamma & 1/2 - \gamma \\ 1/2 - \gamma & 1/2 + \gamma \end{bmatrix}.$$
This defines an $\Omega(\gamma)$-observable POMDP.

Since there are no transitions, the true belief after observations $o_{2:h+t}$ is always $$\belief_{h+t}(a_{1:h+t-1},o_{2:h+t}) = P(\cdot|o_{2:h},b_1) = (1,0).$$ The approximate belief is $$\sbelief_{h+t}(a_{h:h+t-1},o_{h+1:h+t}) \propto \left(\frac{1}{2}(1/2+\gamma)^k (1/2 - \gamma)^{t-k}, \frac{1}{2} (1/2 - \gamma)^k (1/2 + \gamma)^{t-k}\right) := \tilde{b}_t$$ where $k := |\{i \in [t]: o_{h+i} = 0\}|$. If $k \leq t/2$, then $\tilde{b}_t(1) \geq \tilde{b}_t(0)$, so $\sbelief_{h+t}(a_{h:h+t-1},o_{h+1:h+t})_1 \geq 1/2$. Otherwise, $$\sbelief_{h+t}(a_{h:h+t-1},o_{h+1:h+t})_1 = \frac{\tilde{b}_t(1)}{\tilde{b}_t(0) + \tilde{b}_t(1)} \geq \frac{\tilde{b}_t(1)}{2\tilde{b}_t(0)} = \frac{1}{2} \frac{(1/2-\gamma)^k(1/2+\gamma)^{t-k}}{(1/2+\gamma)^k(1/2-\gamma)^{t-k}} = \frac{1}{2}\left(\frac{1/2-\gamma}{1/2+\gamma}\right)^{2k-t}.$$
Assuming $\gamma \leq 0.1$ we have $$\frac{1/2 - \gamma}{1/2 + \gamma} \geq 1 - 4\gamma \geq e^{-8\gamma},$$
so $$\norm{\belief_{h+t}(a_{1:h+t-1},o_{2:h+t}) - \sbelief_{h+t}(a_{h:h+t-1},o_{h+1:h+t})}_1 = \sbelief_{h+t}(a_{h:h+t-1},o_{h+1:h+t})_1 \geq \frac{1}{2}\exp(-8(2k-t)\gamma).$$
Thus, for any $\ep \in (0,1/2)$,
$$\Pr[\norm{\belief_{h+t} - \sbelief_{h+t}}_1 \leq \epsilon] \leq \Pr\left[2k-t \geq \frac{1}{8\gamma}\log \frac{1}{2\epsilon}\right].$$
Setting $\epsilon = e^{-rt}/2$ for some $r>0$, we have $$\Pr[\norm{\belief_{h+t} - \sbelief_{h+t}}_1 \leq e^{-rt}] \leq \Pr\left[k \geq \frac{t}{2} + \frac{rt}{16\gamma}\right].$$
Since $k \sim \text{Bin}(t, 1/2 + \gamma)$, if we take $r = 16\gamma^2$ then the tail bound is $\Pr[k \geq t(1/2 + \gamma)] < 1 - \Omega(1)$. So $$\EE[\norm{\belief_{h+t} - \sbelief_{h+t}}_1] \geq \Omega(e^{-16\gamma^2 t})$$ as desired.  
\end{proof}

We also remark that this yields a lower bound complementing Blackwell and Dubins' ``merging of opinions'' result \cite{blackwell1962merging}. They prove unconditionally (i.e. with no observability assumption) that the predicted observations must asymptotically merge. One might ask whether the observations must (unconditionally) merge exponentially fast, which (together with an observability assumption) would directly imply Theorem~\ref{theorem:stability-informal}. The following corollary shows that this is not the case:

\begin{corollary}\label{cor:blackwell-dubins-converse}
There is a constant $c > 0$ such that the following holds. Let $\epsilon > 0$. There is a POMDP and a policy $\pi$, such that $$\EE_{a_{1:h+t-1},o_{2:h+t} \sim \pi} \norm{\bO_{h+t}^\t \belief_{h+t}(a_{1:h+t-1},o_{2:h+t}) - \bO_{h+t}^\t \sbelief_{h+t}(a_{h:h+t-1},o_{h+1:h+t})}_1 \geq \Omega(\epsilon)$$
where $t = 1/\epsilon^2$ and $h \geq 2$, and $\bO_h$ is the observation matrix of the POMDP at time $h$.
\end{corollary}

\begin{proof}
Take $\gamma = \epsilon$ in Proposition \ref{prop:belief-contraction-lower-bound}, and observe that in the above construction, $\bO_h$ contracts TV-distance by exactly a factor of $\gamma$.
\end{proof}

\bibliographystyle{alpha}
\bibliography{bib}

\appendix

\section{Technical lemmas}

\begin{lemma}\label{lemma:sqrt-contraction}
Let $X$ be a nonnegative random variable with $\EE (X-\EE X)\mathbbm{1}[X \in E] \leq -\eta \EE X$ for some event $E$. Then $\EE \sqrt{X} \leq (1-\eta^2/32)\sqrt{\EE X}$. 
\end{lemma}

\begin{proof}
Without loss of generality assume $\EE X = 1$. Then $\EE(X-1)\mathbbm{1}[X \leq 1] \leq -\eta$. By nonnegativity of $X$, $\Pr[X \leq 1] \geq \eta$. Moreover $$-\eta \geq \EE(X-1)\mathbbm{1}[X \leq 1] \geq 2\EE(\sqrt{X} - 1)\mathbbm{1}[X \leq 1],$$
so $\EE \sqrt{X} \mathbbm{1}[X\leq 1] \leq \Pr(X \leq 1) - \eta/2.$ If $\EE \sqrt{X} \leq 1 - \eta/4$ then we are already done, so suppose that $\EE \sqrt{X} > 1 - \eta/4$. Then
\begin{align}
1 - (\EE \sqrt{X})^2
&= \EE (\sqrt{X} - \EE \sqrt{X})^2 \nonumber\\
&\geq \EE (\sqrt{X} - \EE \sqrt{X})^2 \mathbbm{1}[X \leq 1] \nonumber\\
&\geq (\EE \sqrt{X} \mathbbm{1}[X \leq 1] - \Pr[X \leq 1] \EE \sqrt{X})^2 \tag{Jensen's inequality} \\
&\geq (\eta/4)^2 \nonumber
\end{align}
since $\EE\sqrt{X}\mathbbm{1}[X\leq 1] \leq \Pr[X \leq 1] - \eta/2$ whereas $$\Pr[X\leq 1] \cdot \EE\sqrt{X} \geq \Pr[X\leq 1](1 - \eta/4) \geq \Pr[X\leq 1] - \eta/4.$$
Thus, $$\EE \sqrt{X} \leq \sqrt{1 - \eta^2/16} \leq 1 - \frac{\eta^2}{32}.$$
This proves the lemma.
\end{proof}

\subsection{KL divergence}

Recall that we have defined $f_\text{KL}(x) = x - \log x - 1$ for $x \in (0,\infty)$. The following lemma states several basic properties of $f_\text{KL}$, for which we omit the proof:

\begin{lemma}\label{lemma:f-function}
The function $f_\text{KL}(x)$ is nonnegative and convex on $(0,\infty)$. Moreover, for all $x \in [1/2, 3/2]$, it holds that
$$\frac{1}{4}(x-1)^2 \leq f_\text{KL}(x) \leq (x-1)^2.$$
\end{lemma}

\begin{lemma}\label{lemma:alpha-beta}
Let $\alpha,\beta, d > 0$ such that $f_\text{KL}(1+\alpha) = f_\text{KL}(1-\beta) = d$. If $d \leq 1/16$, then $\alpha,\beta \leq 2\sqrt{d}$.
\end{lemma}

\begin{proof}
From Lemma~\ref{lemma:f-function}, we have that $f_\text{KL}(3/2) \geq 1/16$ and $f_\text{KL}(1/2) \geq 1/16$. Since $f_\text{KL}(1+\alpha) = d \leq 1/16$, and $f_\text{KL}$ is increasing on $(1,\infty)$ and decreasing on $(0,1)$, it follows that $1+\alpha \in [1/2, 3/2]$ and similarly $1-\beta \in [1/2, 3/2]$. Thus, by Lemma~\ref{lemma:f-function}, $\alpha^2/4 \leq d$ and $\beta^2/4 \leq d$, as claimed.
\end{proof}

\begin{lemma}\label{lemma:reverse-pinsker}
  Let $P, Q$ be discrete distributions with \begin{align}
\EE_{y \sim P} \left| f_\text{KL}\left( \frac{Q(y)}{P(y)} \right) - \kld{P}{Q} \right| \leq \frac{1}{4} \cdot \kld{P}{Q}\nonumber.
  \end{align}
  Then $\| P-Q\|_1 \geq \min\left(\frac{1}{4} \sqrt{\kld{P}{Q}}, \frac{1}{10}\right)$.
\end{lemma}
\begin{proof}
  We distinguish two cases. First, suppose that $\kld{P}{Q} \leq 1/16$. In this case, we will show that $\norm{P-Q}_1 \geq \frac{1}{4}\sqrt{\kld{P}{Q}}$. For a real number $x$, define $[x]_- = \max\{-x,0\}$. Since $\kld{P}{Q} \leq 1/16$, Lemma \ref{lemma:alpha-beta} gives that $\alpha,\beta \leq 1/2$. Thus, if $f_\text{KL}(Q(y)/P(y)) \leq \kld{P}{Q}$, we have that $Q(y)/P(y) \in [1/2,3/2]$. It follows from this and Lemma \ref{lemma:f-function} that for all $y \in \supp(P)$,
  \begin{align}
\left[ f_\text{KL}\left( \frac{Q(y)}{P(y)} \right) - \kld{P}{Q} \right]_- \geq \left[ \left( \frac{Q(y)}{P(y)} -1 \right)^2 - \kld{P}{Q} \right]_-\nonumber.
  \end{align}
  (In particular, if the left-hand side above is 0, then $Q(y)/P(y) \not \in [1-\alpha,1+\beta]$, which implies that $(Q(y)/P(y) - 1)^2 > \kld{P}{Q}$, meaning that the right-hand side is 0 also.)
  Next, set
  \begin{align}
\MJ := \left\{ y \in \supp(P): \ \left| \frac{Q(y)}{P(y)} - 1 \right| > \frac{1}{2} \cdot \sqrt{\kld{P}{Q}}\right\}\nonumber.
  \end{align}
  Now assume for the sake of contradiction that $\norm{P-Q}_1 \leq \frac{1}{4}\sqrt{\kld{P}{Q}}$. By this assumption and Markov's inequality,
  \begin{align}
\Pr_{y \sim P}[y \in \MJ] \leq \Pr_{y \sim P} \left[ \left| \frac{Q(y)}{P(y)} - 1 \right| > 2 \| P-Q\|_1 \right] \leq 1/2\nonumber.
  \end{align}
  But then
  \begin{align}
    \EE_{y \sim P} \left| f_\text{KL}\left( \frac{Q(y)}{P(y)} \right) - \kld{P}{Q} \right| \geq & \EE_{y \sim P} \left[ \left[ f_\text{KL}\left( \frac{Q(y)}{P(y)} \right) - \kld{P}{Q} \right]_- \right]\nonumber\\
    \geq & \EE_{y \sim P} \left[ \left[ \left( \frac{Q(y)}{P(y)} -1 \right)^2 - \kld{P}{Q} \right]_- \right]\nonumber\\
    \geq &  \EE_{y \sim P} \left[ \One[y \in \bar\MJ] \cdot \left[ \left( \frac{Q(y)}{P(y)} -1 \right)^2 - \kld{P}{Q} \right]_- \right]\nonumber\\
    \geq & \EE_{y \sim P} \left[ \One[y \in \bar \MJ] \cdot \frac{3}{4} \cdot \kld{P}{Q} \right]\nonumber\\
    \geq & \frac{3}{8} \cdot \kld{P}{Q} \nonumber.
  \end{align}
  This contradicts the assumption of the lemma statement, so in fact $\norm{P-Q}_1 \geq \frac{1}{4}\sqrt{\kld{P}{Q}}$.
  
  In the second case, we suppose that $\kld{P}{Q} \geq 1/16$. In this case, we will show that $\norm{P-Q}_1 \geq 1/10$. Indeed, let $$\mathcal{K} := \left\{y \in \supp(P): f\left(\frac{Q(y)}{P(y)}\right) \leq \frac{1}{2}\kld{P}{Q}\right\}.$$
  By the lemma assumption and Markov's inequality,
  \begin{align*}
     \Pr_{y \sim P}[y \in \mathcal{K}]
     &\leq \Pr_{y \sim P}\left[\left|f\left(\frac{Q(y)}{P(y)}\right) - \kld{P}{Q}\right| \geq \frac{1}{2}\kld{P}{Q}\right] \\
     &\leq 1/2.
  \end{align*}
  Moreover, if $f(Q(y)/P(y)) \geq 1/2\kld{P}{Q} \geq 1/32$, then $|Q(y)/P(y) - 1| \geq 1/5$.
  Thus,
  \begin{align*}
  \norm{P-Q}_1
  &= \EE_{y \sim P}\left|\frac{Q(y)}{P(y)} - 1\right| \\
  &\geq \EE_{y \sim P}\left[ \mathbbm{1}[y \in \bar{\MK}] \cdot \left|\frac{Q(y)}{P(y)} - 1\right|\right] \\
  &\geq \EE_{y \sim P}\left[\mathbbm{1}[y \in \bar{\MK}] \cdot \frac{1}{5}\right] \\
  &\geq 1/10
  \end{align*}
  as claimed.
\end{proof}

\subsection{\Renyi divergence}
Define the following functions:
\begin{align}
  g(t) &= 1/t - 1 \nonumber\\
  h(t) &= 1/t + t - 2\nonumber.
\end{align}
\begin{lemma}
  \label{lem:hg-properties}
  The following statements hold true:
  \begin{align}
    g(t) \leq & 2 \cdot |t-1| \qquad \mbox{for $t \geq 1/2$} \nonumber\\
    h(t) \leq & 2 \cdot (t-1)^2 \qquad \mbox{for $t \in [1/2,3/2]$} \nonumber.
  \end{align}
\end{lemma}

Note that for two probability distributions $P,Q$ over a domain $\MX$, we have that
\begin{align}
\E_{x \sim P}\left[ g \left( \frac{Q(x)}{P(x)} \right)\right] = \E_{x \sim P} \left[h \left( \frac{Q(x)}{P(x)} \right) \right]= \chisq{P}{Q} = \exp(\ren{P}{Q}) -1\nonumber.
\end{align}

\begin{lemma}
  \label{lem:h-chisq}
  Let $P,Q$ be discrete distributions on a set $\MX$ with
  \begin{align}
\E_{x \sim P} \left| h\left( \frac{Q(x)}{P(x)} \right) - \chisq{P}{Q} \right| \leq \frac{1}{4}\cdot\chisq{P}{Q}\nonumber.
  \end{align}
  Suppose that $\chisq{P}{Q} \leq 1/6$. Then $\tvd{P}{Q} \geq \frac{1}{4 \sqrt{2}} \cdot \sqrt{\chisq{P}{Q}}$. 
\end{lemma}
\begin{proof}
  Since $\chisq{P}{Q} \leq 1/6$, if $h(Q(x)/P(x)) \leq \chisq{P}{Q}$, then $Q(x) / P(x) \in [1/2,3/2]$. Then by Lemma \ref{lem:hg-properties}, for all $x \in \MX$,
  \begin{align}
\left[h \left( \frac{Q(x)}{P(x)}\right) - \chisq{P}{Q} \right]_- \geq \left[ 2 \cdot \left( \frac{Q(x)}{P(x)} - 1 \right)^2 - \chisq{P}{Q} \right]_-\nonumber.
  \end{align}
  (In particular, if the left-hand side above is 0, then $Q(x)/P(x) \not \in [1/2,3/2]$, which implies that $2 \cdot (Q(x)/P(x) - 1)^2 > 1/2 \geq \chisq{P}{Q}$, meaning that the right-hand side is 0 also.)

  Next, set
  \begin{align}
\MJ := \left\{ x \in \MX \ : \ \left| \frac{Q(x)}{P(x)} - 1 \right| > \frac{1}{2 \sqrt{2}} \cdot \sqrt{\chisq{P}{Q}} \right\}\nonumber.
  \end{align}
  Now assume for the sake of contradiction that $\tvd{P}{Q} \leq \frac{1}{4 \sqrt 2} \sqrt{\chisq{P}{Q}}$. Then by Markov's inequality,
  \begin{align}
\Pr_{x \sim P} [x \in \MJ] \leq \Pr_{x \sim P } \left[ \left| \frac{Q(x)}{P(x)} - 1 \right| > 2 \tvd{P}{Q} \right] \leq 1/2\nonumber.
  \end{align}
  But then
  \begin{align}
    \E_{x \sim P} \left| h \left( \frac{Q(x)}{P(x)}\right) - \chisq{P}{Q} \right| \geq & \E_{x \sim P } \left[ \left[ h\left( \frac{Q(x)}{P(x)}  \right) - \chisq{P}{Q} \right]_-\right]\nonumber\\
    \geq &  \E_{x \sim P } \left[ \left[ 2 \cdot\left( \frac{Q(x)}{P(x)}  -1\right)^2 - \chisq{P}{Q} \right]_-\right]\nonumber\\
    \geq &  \E_{x \sim P } \left[ \One[x \in \bar \MJ] \cdot \left[ 2 \cdot\left( \frac{Q(x)}{P(x)}  -1\right)^2 - \chisq{P}{Q} \right]_-\right]\nonumber\\
    \geq & \E_{x \sim P} \left[ \One[x \in \bar \MJ] \cdot 3/4 \cdot \chisq{P}{Q} \right]\nonumber\\
    \geq & \frac{3}{8} \cdot \chisq{P}{Q}\nonumber,
  \end{align}
  which contradicts the assumption made in the lemma statement.
\end{proof}

\begin{lemma}
  \label{lem:g-chisq}
  Let $P,Q$ be discrete distributions on a set $\MX$ with
  \begin{align}
\E_{x \sim P} \left| g\left( \frac{Q(x)}{P(x)} \right) - \chisq{P}{Q} \right| \leq \frac{1}{10}\cdot\chisq{P}{Q}\nonumber.
  \end{align}
  Then $\tvd{P}{Q} \geq \min \left\{ \frac 18 \cdot {\chisq{P}{Q}}, \frac 17\right\}$. 
\end{lemma}
\begin{proof}
  We first consider the case that $\chisq{P}{Q} \leq 1/2$. If $g(Q(x)/P(x)) \leq \chisq{P}{Q}$, then $Q(x)/P(x) \geq 2/3 \geq 1/2$. Then by Lemma \ref{lem:hg-properties}, for all $x \in \MX$,
  \begin{align}
\left[ g\left( \frac{Q(x)}{P(x)}\right) - \chisq{P}{Q} \right]_- \geq \left[ 2 \cdot \left| \frac{Q(x)}{P(x)} -  1 \right| - \chisq{P}{Q} \right]_- .\nonumber
  \end{align}
  (If particular, if the left-hand side is 0, then $Q(x)/P(x) \leq 2/3$, which implies that $2 \cdot |Q(x)/P(x) - 1| \geq 2/3 > \chisq{P}{Q}$, meaning that the right-hand side is 0 also.)

  Next, set
  \begin{align}
\MJ := \left\{ x \in \MX : \left| \frac{Q(x)}{P(x)} - 1 \right| > \frac{1}{4} \cdot \chisq{P}{Q} \right\}.\nonumber
  \end{align}
  Assume for the sake of contradiction that $\tvd{P}{Q} \leq \frac{1}{8} \cdot \chisq{P}{Q}$. Then by Markov's inequality,
  \begin{align}
\Pr_{x \sim P} [x \in \MJ] \leq \Pr_{x \sim P} \left[ \left| \frac{Q(x)}{P(x)} - 1 \right| > 2 \tvd{P}{Q} \right] \leq 1/2\nonumber.
  \end{align}
  But then
  \begin{align}
    \E_{x \sim P} \left| g \left( \frac{Q(x)}{P(x)}\right) - \chisq{P}{Q} \right| \geq & \E_{x \sim P}\left[ \left[ g \left( \frac{Q(x)}{P(x)}\right) - \chisq{P}{Q} \right]_-\right]\nonumber\\
    \geq & \E_{x \sim P} \left[\left[ 2 \cdot \left| \frac{Q(x)}{P(x)} -  1 \right| - \chisq{P}{Q} \right]_-\right]\nonumber\\
    \geq &  \E_{x \sim P} \left[\One[x \in \bar\MJ] \cdot \left[ 2 \cdot \left| \frac{Q(x)}{P(x)} -  1 \right| - \chisq{P}{Q} \right]_-\right]\nonumber\\
    \geq & \E_{x \sim P} \left[ \One[x \in \bar \MJ] \cdot \frac 14 \cdot \chisq{P}{Q} \right]\nonumber\\
    \geq & \frac 18 \cdot \chisq{P}{Q}\nonumber.
  \end{align}
  This contradicts the assumption in the lemma statement, giving us that $\tvd{P}{Q} \geq \frac 18 \cdot \chisq{P}{Q}$ in this case.

  We next consider the case that $\chisq{P}{Q} > 1/2$. Let
  \begin{align}
\MK := \left\{ x \in \MX : g \left( \frac{Q(x)}{P(x)} \right) \leq \frac 12 \cdot \chisq{P}{Q} \right\}.\nonumber
  \end{align}
  By the lemma assumption and Markov's inequality,
  \begin{align}
    \Pr_{x \sim P} \left[ x \in \MK \right] \leq & \Pr_{x \sim P} \left[ \left| g \left( \frac{Q(x)}{P(x)} \right) - \chisq{P}{Q} \right| \geq \frac 12 \chisq{P}{Q} \right]\nonumber\\
    \leq & 1/5\nonumber.
  \end{align}
  Moreover, if $g(Q(x)/P(x)) > \frac 12 \cdot \chisq{P}{Q} \geq 1/4$, then $|Q(x)/P(x) - 1| > 1/5$. Then
  \begin{align}
    \tvd{P}{Q} =& \E_{x \sim P} \left| \frac{Q(x)}{P(x)} - 1 \right| \nonumber\\
    \geq & \E_{x \sim P} \left[ \One[x \in \bar \MK] \cdot \left| \frac{Q(x)}{P(x)} - 1 \right| \right]\nonumber\\
    \geq & \E_{x \sim P} \left[ \One[x \in \bar \MK] \cdot \frac 15 \right]\nonumber\\
    \geq & 4/5 \cdot 1/5 \geq 1/7\nonumber
  \end{align}
  as desired.
\end{proof}


\begin{lemma}
  \label{lem:24-alpha}
  Consider the function
  \begin{align}
    f_1(x) = \left( \sqrt{x^2+1} - 1 \right)^{1/2}, 
    \nonumber
  \end{align}
  defined for $x \geq 0$.
  Then for any $\alpha \in [0,1/2]$, and $x \geq 0$,
  \begin{align}
    f_1((1-\alpha)x) \leq (1-\alpha/8) \cdot f_1(x)
    \nonumber.
  \end{align}
\end{lemma}
\begin{proof}
  For $0 \leq y \leq 1$, we have that
  \begin{align}
\sqrt{y+1} - \sqrt{(1-\alpha)y+1} \geq \alpha y \cdot \frac{1}{2 \sqrt{y+1}} \geq \frac{\alpha y}{2 \sqrt 2}\nonumber
  \end{align}
  and $\sqrt{y+1} - 1 \leq y/2$. Thus
  \begin{align}
\sqrt{(1-\alpha)y+1} - 1 \leq \sqrt{y+1}-1 - \frac{\alpha y}{2\sqrt 2} \leq \left(1 - \alpha/\sqrt 2 \right) \cdot (\sqrt{y+1}-1)\nonumber.
  \end{align}
  Furthermore, for $y \geq 1$, we have
  \begin{align}
    \sqrt{(1-\alpha)y+1} - 1 \leq & \sqrt{(1-\alpha/2)(y+1)} - 1 \nonumber\\
    \leq & (1-\alpha/4) \cdot \sqrt{y+1}-1\nonumber\\
    \leq & (1-\alpha/4) \cdot (\sqrt{y+1}-1)\nonumber.
  \end{align}
  Therefore,
  \begin{align}
    f_1((1-\alpha)x) =& \left( \sqrt{(1-\alpha)^2x^2 + 1} - 1 \right)^{1/2} \nonumber\\
    \leq & \left( \sqrt{(1-\alpha) x^2 + 1} - 1 \right)^{1/2}\nonumber\\
    \leq & \left( (1-\alpha/4) \cdot (\sqrt{x^2+1}-1) \right)^{1/2}\nonumber\\
    \leq & (1-\alpha/8) \cdot f_1(x)\nonumber
  \end{align}
  as desired.
  \end{proof}
  
The following lemma is an immediate corollary of Lemma \ref{lem:24-alpha}.
\begin{lemma}
  \label{lem:24-alpha-cor}
Define $f_1(x) = \left( \sqrt{x^2+1}-1 \right)^{1/2}$. Suppose that $X$ is a non-negative valued random variable and $\E[X] \leq (1-\eta) a$, for some $\eta \in [0,1/2]$. Then $\E[f_1(X)] \leq (1-\eta/8) \cdot f_1(a)$.
\end{lemma}
\begin{proof}
  It is straightforward to see that $f_1$ is concave and monotonically increasing for $x \geq 0$. Thus, by Jensen's inequality, we have
  \begin{align}
\E[f_1(X)] \leq f_1(\E[X]) \leq f_1((1-\eta)a) \leq (1-\eta/8) f_1(a),\nonumber
  \end{align}
  where the final inequality uses Lemma \ref{lem:24-alpha}.
\end{proof}

\section{Examples}

\begin{example}[Simple observation models satisfying observability]\label{ex:simple-observable}
Let $\gamma > 0$. Here are two $\gamma$-observable models:
\begin{enumerate}
    \item \label{it:obs-null} Let $\MS$ be any state space, and let $\MO = \MS \sqcup \{\perp\}$. At state $i \in \MS$, say that the observation is $i$ with probability $\gamma$, and a ``null observation'' $\perp$ otherwise. This describes an observation matrix $\bO$, which we claim is $\gamma$-observable. Indeed, for any $b \in \Delta^\MS$, we have $(\bO^\t b)(\perp) = 1/2$, and $(\bO^\t b)(i) = \gamma b(i)$ for any $i \in \MS$. Thus, for $b,b' \in \MS$, we have $$\norm{\bO^\t b - \bO^\t b'}_1 = \gamma \norm{b - b'}_1.$$
    
    \item \label{it:gamma-identity} Let $\MS$ be any state space, and let $\MO = \MS$. At state $i \in \MS$, say that the observation is $i$ with probability $\gamma$, and otherwise a uniformly random state. We claim that the resulting observation matrix $\bO$ is $\gamma$-observable. Indeed, for any $b \in \Delta^\MS$ and any $i \in \MS$, we have $(\bO^\t b)(i) = \gamma b(i) + \frac{1-\gamma}{S}$. Thus, for $b,b' \in \MS$, we have that $$\norm{\bO^\t b - \bO^\t b'}_1 = \gamma \norm{b-b'}_1.$$
\end{enumerate}
\end{example}

\begin{example}[No sub-exponential size net over achievable beliefs]\label{ex:large-net-needed}
We define an observable POMDP as follows. Fix $m \in \BN$. The state space is $[m]$ with uniform initial state distribution. The horizon length is $m$. The observation space is $[m] \cup \{\perp\}$, and for any state $i$, the observation is $i$ with probability $1/2$ and $\perp$ otherwise; this satisfies observability with $\gamma = 1/2$ (Example~\ref{ex:simple-observable}). The action space is $[m] \times [m]$; action $(i,j)$ is the deterministic transition such that if the current state is $i$, then the next state will be $j$, and otherwise the state remains fixed at its current value. Then for any belief state $P \in \Delta^{[m]}$ such that $mP(x)$ is an integer for all $x$, we can achieve $P$ by the appropriate action/observation sequence where the observations are all $\perp$ (this has positive probability) and the actions move mass as necessary. Thus, for e.g. $\epsilon = 1/2$, any net on $\Delta^{[m]}$ needs size $\exp(\Omega(m))$ to cover all reachable beliefs with $\epsilon$-TV-balls.
\end{example}

\begin{example}[Bayes update can sometimes increase KL-divergence]\label{ex:divergence-increase}
Let $\MS = \MO = \{1,2\}$ and let $\epsilon>0$. Define the observation matrix by $\bO_h(o|s) = (1-\epsilon)\mathbbm{1}[o=s] + \epsilon\mathbbm{1}[o\neq s]$. Take $b = (1-\epsilon^2,\epsilon^2)$ and $b' = (1/2,1/2)$. Then $\kld{b}{b'} \leq \log(2)$. But $$B_h(b;2) = \frac{1}{\epsilon+2\epsilon^2 - \epsilon^3}(\epsilon-\epsilon^3, \epsilon^2-\epsilon^3) = (1 - O(\epsilon), O(\epsilon))$$ and $$B_h(b';2) = (\epsilon, 1-\epsilon),$$ so $\kld{B_h(b;2)}{B_h(b';2)} = \Omega(\log(1/\epsilon))$, which is greater than $\log(2)$ for sufficiently small $\epsilon$. Of course, the event $o=2$ is unlikely under the state distribution $b$, and if $o=1$ then the Bayes update will decrease the KL-divergence significantly.
\end{example}

\begin{example}[Expected KL-divergence may not decrease at linear rate in one step]\label{ex:no-linear-rate}
Define a POMDP with state space and observation space $\MS = \MO = \{0,1\}$. Let $b = (1,0)$ and $b' = (1-\epsilon, \epsilon)$, and let the observations for state $0$ be distributed $(1/2+\gamma, 1/2-\gamma)$, and for state $1$ be distributed $(1/2-\gamma,1/2+\gamma)$. Define $f(x) = \log(1+x)$. Then $$\kld{b}{b'} = \log \frac{1}{1-\epsilon} = f\left(\frac{\epsilon}{1-\epsilon}\right).$$
Similarly, $$\kld{B_h(b,0)}{B_h(b',0)} = f\left(\frac{1/2-\gamma}{1/2+\gamma} \cdot \frac{\epsilon}{1-\epsilon}\right)$$ and $$\kld{B_h(b,1)}{B_h(b',1)} = f\left(\frac{1/2+\gamma}{1/2-\gamma} \cdot \frac{\epsilon}{1-\epsilon}\right),$$
so that $$\EE_{y \sim \bO_h^\t b} \kld{B_h(b,y)}{B_h(b',y)} = (1/2 + \gamma) f\left(\frac{1/2-\gamma}{1/2+\gamma} \cdot \frac{\epsilon}{1-\epsilon}\right) + (1/2 - \gamma)f\left(\frac{1/2+\gamma}{1/2-\gamma} \cdot \frac{\epsilon}{1-\epsilon}\right).$$
But $f(x) = x - \frac{1}{2}x^2 + O(x^3)$, so $$\kld{b}{b'} = \frac{\epsilon}{1-\epsilon} - \frac{1}{2}\left(\frac{\epsilon}{1-\epsilon}\right)^2 + O(\epsilon^3),$$ whereas $$\EE_{y \sim \bO_h^\t b} \kld{B_h(b,y)}{B_h(b',y)} = \frac{\epsilon}{1-\epsilon} - \frac{(1+O(\gamma^2))}{2}\left(\frac{\epsilon}{1-\epsilon}\right)^2 + O(\epsilon^3).$$
Thus, the decrease is $$\kld{b}{b'} - \EE_{y \sim \bO_h^\t b} \kld{B_h(b,y)}{B_h(b',y)} = O(\gamma^2 \epsilon^2) + O(\epsilon^3)$$
which is quadratic in $\kld{b}{b'}$.
\end{example}

\section{Succinct Descriptions of Policies}\label{app:succinct-policies}

In this section we extend the result of \cite{papadimitriou1987complexity} that optimal policies cannot be succinctly described unless $\PSPACE = \Sigma_2^P$ in two ways: first, we extend it to \emph{approximately}-optimal policies, and second, we relax the definition of succinctness (from polynomial succinctness to quasi-polynomial succinctness). 

\begin{definition}\label{def:succinct}
Let $\epsilon > 0$. For a function $f: \BN \to \BN$, we say that $\epsilon$-optimal POMDP policies have $f$-succinct descriptions if there is a polynomial-time Turing machine $T$ with the following property: for any POMDP $M$ with all rewards in $[0,1]$ and size at most $n$, there is some string $d(M)$ of size $|d(M)| \leq f(n)$, and some policy $(\pi_h)_{h \in [H]}$ such that:
\begin{itemize}
\item $V^\pi(M) \geq V^*(M) - \epsilon$
\item $T(d(M), (a_{1:h-1},o_{2:h})) = \pi(a_{1:h-1},o_{2:h})$ for all $h \in [H]$ and $(a_{1:h-1},o_{2:h}) \in (\MA \times \MO)^{h-1}$.
\end{itemize}
We can make an analogous definition for any family of POMDPs $\mathcal{M}$ (in particular, the policy $M$ is constrained to lie in $\mathcal{M}$). 
\end{definition}

The guarantee of Theorem \ref{thm:main-intro} (which uses Algorithm \ref{alg:apvi} to compute a near-optimal policy)  shows that for any constant $\gamma>0$, the family of $\gamma$-observable POMDPs with $\max\{ H, S, A, O \} \leq n$ has $n^{O(\log(n/\ep))}$-succinct descriptions for computing $\ep$-optimal policies. Below we will show that this guarantee does not extend to the family of all POMDPs, unless the exponential hierarchy collapses. First we need the following result, which shows that approximating the optimal value of a POMDP is $\PSPACE$-complete.

\begin{theorem}[\cite{lusena2001nonapproximability}]\label{theorem:lusena}
Let $\epsilon > 0$, and let $L \in \PSPACE$ be a language. There is a polynomial-time computable function $g$ and a polynomial-time computable polynomial $p$ such that for any $n \in \BN$ and $x \in \{0,1\}^n$, the output $g(x)$ is a POMDP $M$ satisfying
\begin{itemize}
    \item If $x \in L$ then the value of $M$ is at least $p(n)$
    \item If $x \not \in L$ then the value of $M$ is strictly less than $p(n) - \epsilon$
    \item All rewards in $M$ lie in $[0,1]$.
\end{itemize}
\end{theorem}

\begin{proof}
This theorem follows by inspecting the proof of Theorem~4.11 in \cite{lusena2001nonapproximability}.
\end{proof}

\begin{theorem}\label{theorem:succinct-impossibility}
Fix any $\epsilon, c > 0$. Define function $f(n) = 2^{\log^c(n)}$ and suppose that $\epsilon$-optimal POMDP policies have $f$-succinct descriptions. Then the exponential hierarchy collapses to level $2$.
\end{theorem}

\begin{proof}
Let $L$ be any \PSPACE-complete language (e.g. TQBF). We claim that there is a polynomial-time probabilistic Turing machine $T^V$ such that for any $n \in \BN$ and $x \in \{0,1\}^n$,
\begin{itemize}
    \item If $x \in L$, then there exists $z$ with $|z| \leq f(n)$ such that $\Pr[T^V(x,z) = 1] \geq 2/3$
    \item If $x \not \in L$, then for every $z$ with $|z| \leq f(n)$, it holds that $\Pr[T^V(x,z) = 1] \leq 1/3$.
\end{itemize}
Indeed, let $g$ and $p$ be the function and polynomial guaranteed by Theorem~\ref{theorem:lusena} with gap $3\epsilon$. Let $q$ be a polynomial such that $q(n)$ bounds the size of the POMDP $g(x)$ for $x \in \{0,1\}^n$. Let $T$ be the Turing machine guaranteed by Definition~\ref{def:succinct}. Define $T^V$ as follows. On input $(x,z)$, compute $M = g(x)$. Take $N = 100q(n)/\epsilon^2$, and simulate $N$ trajectories and their rewards $r_1,\dots,r_N$ by following the policy described by $T(z, \cdot)$ (if at any point this outputs an invalid action, then terminate and output $0$). Let $\bar{r} = \frac{1}{N}\sum_{i=1}^N r_i$. If $\bar{r} \geq p(n) - 2\epsilon$ then output $1$; otherwise output $0$.

We prove that $T^V$ has the desired properties. Suppose $x \in L$. Then $V^*(M) \geq p(n)$, so the policy $\pi$ described by $d(M)$ satisfies $V^\pi(M) \geq p(n) - \epsilon$. Take $z = d(M)$, which by assumption has $|z| \leq f(n)$, and consider the execution of $T^V(x,z)$. Then $r_1,\dots,r_N$ are independent random samples from the reward distribution of policy $\pi$ on $M$, which has expectation $V^\pi$ and is always bounded between $0$ and $q(n)$. Thus, by Hoeffding's inequality, $$\Pr[\bar{r} - V^\pi < -\epsilon] \leq \exp(-2\epsilon^2 N / q(n)) \leq 1/3.$$
In the converse event, we have $\bar{r} \geq V^\pi - \epsilon \geq p(n) - 2\epsilon$. Thus, $\Pr[T^V(x,z) = 1] \geq 2/3$.

Conversely, suppose that $x \not \in L$. Then $V^*(M) \leq p(n) - 3\epsilon$. Take any $z$ with $|z| \leq f(n)$. Without loss of generality, $T(z,\cdot)$ describes a valid policy $\pi$ (whenever the output is invalid we could take an arbitrary fixed action, and this only increases the rewards). As before, $r_1,\dots,r_N$ are independent random samples from distribution bounded between $0$ and $q(n)$, with mean $V^\pi$, so by Hoeffding's inequality, $$\Pr[\bar{r} - V^\pi \geq \epsilon] \leq \exp(-2\epsilon^2 N/q(n)) \leq 1/3.$$
In the converse event, we have $\bar{r} < V^\pi + \epsilon \leq p(n) - 2\epsilon$. So $\Pr[T^V(x,z) = 0] \geq 2/3$.

This proves the claim that $\PSPACE \subseteq \MerlinArthur(f)$, i.e. Merlin-Arthur with advice length $f(n)$. Since $\coNP \subseteq \PSPACE$ and $\MerlinArthur(f) \subseteq \AM[f,3] \subseteq \AM(f^{O(1)})$ (see Theorem~2.3 in \cite{goldreich2002interactive}), we get $\coNP \subseteq \AM(2^{\log^k(n)})$ for some constant $k$. By Theorem~3.7 in \cite{pavan2007polylogarithmic}, it holds that $\EH \subset \AM_{\exp}$, which implies that the exponential hierarchy collapses to level $2$ since $\AM_{\exp} \subset \Pi_2^{\exp}$ (see the proof of \cite[Proposition 1]{babai1988arthurmerlin}, which extends in a straightforward manner to the parametrized version). 
\end{proof}

\end{document}